\documentclass[12pt, draftclsnofoot, onecolumn]{IEEEtran}
\usepackage[utf8]{inputenc}

\usepackage{amsmath}
\usepackage{amssymb}
\usepackage{amsfonts}
\usepackage{cite}

\DeclareMathOperator*{\argmax}{arg\,max}

\newcommand{\Rmnum}[1]{\expandafter\@slowromancap\romannumeral #1@}

\usepackage{booktabs}
\usepackage{bbm}
\usepackage{mathrsfs}
\usepackage[caption=false]{subfig}

\usepackage{graphicx}
\usepackage{epsfig}
\usepackage[numbers,sort&compress]{natbib}
\usepackage{xcolor}
\usepackage{color}
\usepackage{enumerate}
\usepackage{extarrows}
\usepackage[ruled,linesnumbered,lined]{algorithm2e}  
\usepackage{algorithmic}
\usepackage{tensor}
\usepackage[colorlinks,linkcolor=blue,anchorcolor=blue,citecolor=blue,urlcolor=black]{hyperref}
\usepackage{url}

\usepackage{xcolor,soul}

\begin{document}

\title{Neuromorphic Wireless Split Computing with Multi-Level Spikes}
\author{Dengyu Wu, \IEEEmembership{Member,~IEEE}, Jiechen Chen,  \IEEEmembership{Member,~IEEE},~ \\Bipin Rajendran,~\IEEEmembership{Senior Member,~IEEE},~ H. Vincent Poor,  \IEEEmembership{Life Fellow,~IEEE},~\\Osvaldo Simeone,~\IEEEmembership{Fellow,~IEEE}
\thanks{D. Wu and B. Rajendran are with the King's Laboratory for Intelligent Computing (KLIC) lab within the Centre for Intelligent Information Processing Systems (CIIPS) at the Department of Engineering, King’s College London, London, WC2R 2LS, UK (email:\{dengyu.wu, bipin.rajendran\}@kcl.ac.uk). J. Chen and O. Simeone are with the King’s Communications, Learning and Information Processing (KCLIP) lab within the CIIPS at the Department of Engineering, King’s College London, London, WC2R 2LS, UK (email:\{jiechen.chen, osvaldo.simeone\}@kcl.ac.uk). H. Vincent Poor is with the Department of Electrical and Computer Engineering, Princeton University, Princeton, NJ 08544 USA (e-mail:poor@princeton.edu).\\
This work was supported by the European Union’s Horizon Europe project CENTRIC (101096379),  by~an Open Fellowship of the EPSRC (EP/W024101/1), by the EPSRC project (EP/X011852/1), and by the U.S. National Science Foundation under Grant ECCS-2335876.
 }
\vspace*{-0.9cm}
}

\maketitle

\IEEEpeerreviewmaketitle
\newtheorem{definition}{\underline{Definition}}[section]
\newtheorem{fact}{Fact}
\newtheorem{assumption}{Assumption}
\newtheorem{theorem}{Theorem}
\newtheorem{lemma}{\underline{Lemma}}[section]
\newtheorem{proposition}{\underline{Proposition}}[section]
\newtheorem{corollary}[proposition]{\underline{Corollary}}
\newtheorem{example}{\underline{Example}}[section]
\newtheorem{remark}{\underline{Remark}}[section]
\newcommand{\mv}[1]{\mbox{\boldmath{$ #1 $}}}
\newcommand{\mb}[1]{\mathbb{#1}}
\newcommand{\Myfrac}[2]{\ensuremath{#1\mathord{\left/\right.\kern-\nulldelimiterspace}#2}}
\newcommand\Perms[2]{\tensor[^{#2}]P{_{#1}}}
\newcommand{\note}[1]{[\textcolor{red}{\textit{#1}}]}

\begin{abstract}
Inspired by biological processes, neuromorphic computing leverages spiking neural networks (SNNs) to perform inference tasks, offering significant efficiency gains for workloads involving sequential data. Recent advances in hardware and software have shown that embedding a small payload within each spike exchanged between spiking neurons can enhance inference accuracy without increasing energy consumption. To scale neuromorphic computing to larger workloads, split computing — where an SNN is partitioned across two devices — is a promising solution. In such architectures, the device hosting the initial layers must transmit information about the spikes generated by its output neurons to the second device. This establishes a trade-off between the benefits of multi-level spikes, which carry additional payload information, and the communication resources required for transmitting extra bits between devices. This paper presents the first comprehensive study of a neuromorphic wireless split computing architecture that employs multi-level SNNs. We propose digital and analog modulation schemes for an orthogonal frequency division multiplexing (OFDM) radio interface to enable efficient communication. Simulation and experimental results using software-defined radios reveal performance improvements achieved by multi-level SNN models and provide insights into the optimal payload size as a function of the connection quality between the transmitter and receiver.

\end{abstract}


\section{Introduction}

\subsection{Context and Motivation}
Current learning algorithms, computing primitives, and hardware platforms such as GPUs are widely expected to soon fall short in supporting scalable, energy-efficient artificial intelligence (AI) models, especially for edge deployments \cite{bourzac2024fixing}. This motivates the ongoing exploration of alternative computing paradigms, including in-memory computing \cite{sebastian2020memory}, neuromorphic computing \cite{liu2014event,simeone2019learning}, and quantum computing \cite{biamonte2017quantum,simeone2022introduction}. Advances in computing technologies are bound to affect a range of fields from the sciences \cite{baker_hassabis_jumper_2024} to engineering \cite{bartolozzi2022embodied}. This work studies some of  the implications of  the emergence of neurormorphic computing for telecommunications  engineering \cite{skatchkovsky2020end, chen2023neuromorphic, chen2022neuromorphic, 10682971, 10694106, liu2024energy, gupta2024spikingrx, martini2022lossless}.

As communication networks become increasingly softwarized \cite{oran2020whitepaper}, spiking neural networks (SNNs) present a promising option as co-processors for wireless transmitters and receivers, as explored in \cite{gupta2024spikingrx, 10694106}. Neuromorphic computing, therefore, can play an important role in enabling advanced communication functionalities. Conversely, communication networks can support the development of distributed computing architectures grounded in neuromorphic principles. In these architectures, communication protocols must be tailored to the unique nature of information exchanged between SNN neurons. Unlike conventional multi-bit clocked messages, spiking neurons encode and transmit information through the  timing of individual spikes.

\begin{figure}[ht!]
    \centering
    \includegraphics[width=\linewidth]{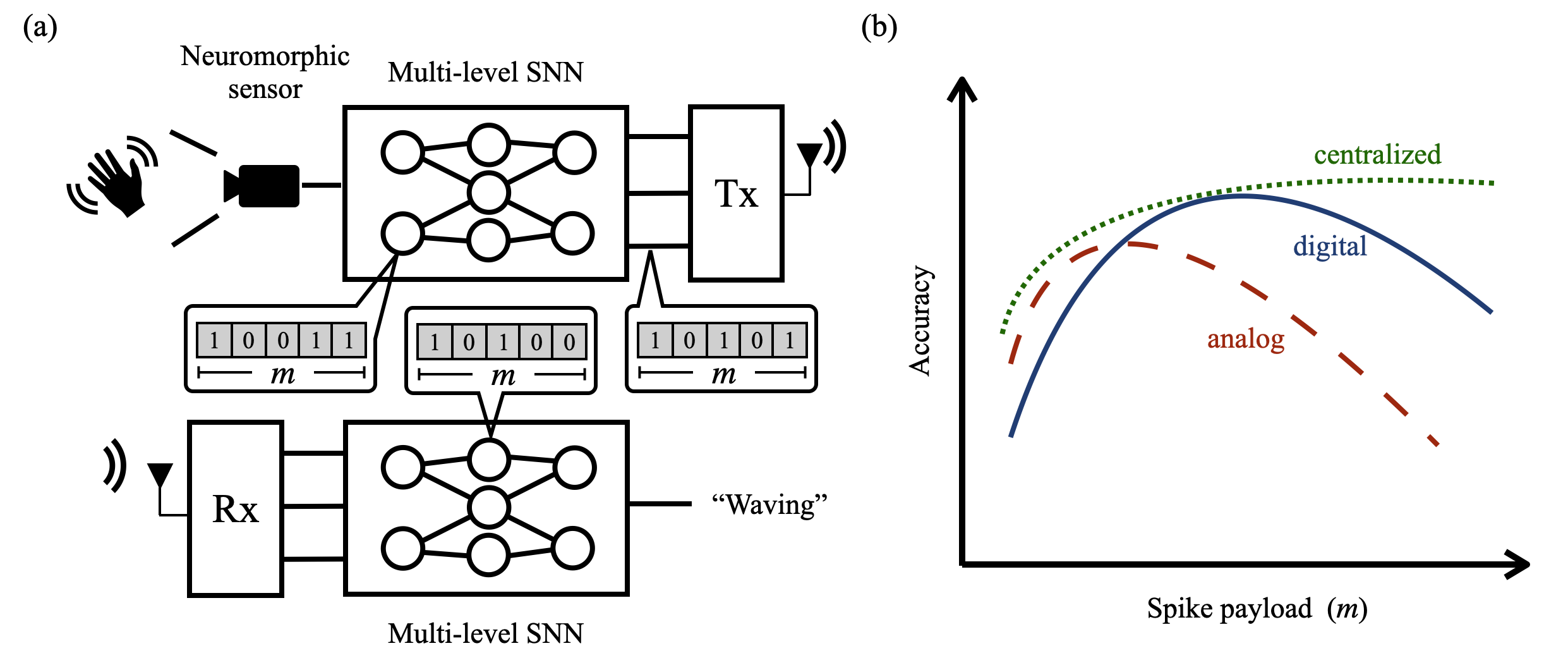}
    \caption{(a) Neuromorphic wireless split computing architecture based on multi-level SNNs: Spikes exchanged between a transmitter and a receiver over a wireless channel include a payload of $m$ bits. (b) While the accuracy of a centralized implementation increases monotonically with the spike payload $m$  \cite{shrestha2024efficient,theilman2024spiking}, in the presence of communication constraints there is generally an optimized value of $m$ that balances the informativeness of each spike with the reduced accuracy of higher-rate transmission.   }
    \label{fig: system}
\end{figure}

While SNNs can reduce the energy consumption for certain workloads \cite{wu2022little}, large-scale tasks requiring deeper SNN architectures may still prove too demanding in terms of energy and memory for mobile devices \cite{davies2021advancing}. In these settings, split computing --- where the computational workload is distributed across multiple devices --- is promising solution. In particular, in \cite{di2024ec}, the authors demonstrated that splitting a deep SNN architecture across multiple edge devices reduced inference latency by 60.7\% and the overall energy consumption per device by 27.7\%. However, partitioning an SNN across multiple devices requires the devices to share timing information to maintain the integrity of the neural computations \cite{skatchkovsky2020end, skatchkovsky2020federated, chen2023neuromorphic, 10682971, borsos2022resilience, martini2022lossless, venkatesha2021federated}.

As shown in Fig. \ref{fig: system}(a), in this paper, we focus on a basic distributed computing architecture \cite{matsubara2020split} consisting of an SNN split between two devices, which  are connected over a wireless channel. The transmitter-side SNN processes sequential data captured by a neuromorphic sensor, such as an event-driven  camera \cite{event, event2,event3, event4}. SNNs can natively process event-driven data via spiking neurons. The receiver-side SNN uses the received radio signal to produce a final inference decision. For example, in the set-up shown in Fig. \ref{fig: system}, the transmitter's sensor observes hand gestures, which are estimated at the receiver side. 

Conventional SNNs represent information solely in the timing of  spikes. However, digital neuromorphic chips, such as Intel's Loihi 2 supports multi-level, or graded, spikes with minimal additional energy cost \cite{shrestha2024efficient,theilman2024spiking} (see Sec. II.B for further details). Multi-level spikes encode information both in the timing of the spikes and in their amplitude. There is evidence that biological brains may also leverage spike amplitude variability to encode additional information \cite{tchumatchenko_magnusson_2014}.   As illustrated in Fig. 1(a), multi-level spikes are assigned a payload of $m$ bits, while conventional spike carry $m=0$ additional bits of information. SNNs with multi-level spikes have been shown to improve the accuracy of conventional SNN deployments, particularly when the number of timesteps available for inference is limited \cite{10472977,hao2024lm,xiao2024multi}.

In a split computing architecture, the introduction of multi-level spikes creates the challenge of transmitting a larger amount of information per spike on the wireless interface. As illustrated in Fig. \ref{fig: system}(b), while in a centralized implementation, larger values of payload size $m$ are generally beneficial in terms of inference accuracy, in a split computing system, an excessively large payload size can cause a performance degradation due to the lower fidelity of higher-rate transmissions on wireless channels. This work addresses this inherent tension by investigating the design of both analog and digital transmission schemes for neuromorphic wireless split computing systems with multi-level spikes.

\subsection{Related Work}
\emph{Neuromorphic wireless split computing}: Neuromorphic wireless split computing  was first studied in \cite{skatchkovsky2020end}, in which single-link neuromorphic sensing and computing were integrated with ultra-wideband (UWB)   transmission to enable edge-based remote inference. This work was then extended in \cite{chen2023neuromorphic} to a multi-device scenario with frequency-selective channels, demonstrating IR transmission’s compatibility in multi-device environments. In \cite{10682971}, wake-up radios were incorporated into the system to further reduce the overall energy consumption of the system.

The work reported in \cite{borsos2022resilience}  analyzed how spike losses affect the inference accuracy and total neural activity when considering a distributed wireless SNN implementation. Another reference \cite{liu2024energy} studied a distributed system of edge nodes, each containing a subset of spiking neurons, that communicate with an access point via wireless channels using frequency division multiple access (FDMA) by allocating different frequency bands to different nodes. 

A neuromorphic integrated sensing and communications system was studied in \cite{chen2022neuromorphic}, in which an SNN was deployed at the receiver to decode the transmitted information and detect the possible presence of a target simultaneously.

There have been also several reported prototypes for neuromorphic split computing. The transmission model in \cite{racz2022full} utilized neuromorphic principles, implemented on Intel's Loihi chip combined with software-defined radio (SDR) hardware, to build a full-stack neuromorphic wireless communication system that considers both orthogonal frequency division multiplexing (OFDM) and UWB transmission. Another work \cite{lee2024asynchronous} experimentally demonstrated a communication approach for large-scale wireless asynchronous microsensor networks, enabling the transmission of binary events from thousands of local nodes with high spectral efficiency and low error rates.

\emph{Multi-level SNNs}: A few studies have contributed to advances in multi-level SNNs in centralized implementations. For example, reference \cite{xiao2024multi} proposed a multi-bit transmission mechanism that expands spike representation from a single bit to multiple bits,  enriching the information content per spike. In \cite{guo2024ternary}, a ternary spiking neuron was introduced to increase information capacity while retaining event-driven, addition-only processing advantages. Additionally, reference \cite{luo2024integer} designed a spiking neuron that activates integer values during training and maintains spike-driven behavior by extending virtual time-steps during inference for object detection tasks.

\subsection{Main Contributions}

This paper investigates for the first time neuromorphic wireless split computing with  multi-level SNNs. Unlike conventional neural networks, SNNs are sequential models, processing and transmitting information over time. Furthermore, they use timing information for inter-neuron communication, producing temporally sparse signals. This is fundamentally different from the dense and continuous representations of inter-neuron signals in artificial neural networks (ANNs). For these reasons, SNNs require separate designs and evaluations as compared to conventional neural networks \cite{chen2023neuromorphic, 10682971}.

Previous works \cite{10682971, skatchkovsky2020end, chen2023neuromorphic, chen2022neuromorphic}, which focused on conventional SNNs, adopted a UWB interface due to its low power consumption and compatibility with spike-based transmission. In contrast,  in this paper we adopt the standard OFDM interface, which provides a more flexible modulation scheme to accommodate multi-level spikes and is more widely applicable and available. In particular,  OFDM facilitates prototyping using conventional SDR platforms, such as the Universal Software Radio Peripheral (USRP) \cite{marwanto2009experimental}. We design and evaluate both digital and analog modulation schemes, which are tested via simulation and via an experimental platform. 

Overall, the main contributions of this paper are summarized as follows.
\begin{itemize}
    \item We study for the first time a neuromorphic wireless split computing architecture based on multi-level SNNs. Unlike conventional SNNs with binary spikes, multi-level SNNs are able to process richer information by assigning a multi-bit payload to each spike.
    \item We detail digital and analog transmission schemes that leverage the sparsity of inter-neuron signals in SNNs, and adapt to the dynamic output of spikes produced over time. The proposed digital modulation scheme is based on the  address-event representation (AER) of multi-level spikes \cite{shrestha2024efficient,theilman2024spiking}. In this implementation, spike addresses and payloads are channel-encoded and modulated on OFDM symbols. If the number of information bits exceeds the available capacity -- which is  more likely to occur for a larger value of the payload size $m$ -- spikes are dropped, causing a potential decrease in accuracy. Upon channel decoding, the transmitted spikes are reconstructed at the receiver and fed to the receiver-side SNN to produce the final inference decision. 
    \item We also detail an analog implementation whereby each output neuron of the transmitter-side SNN is assigned to a fixed subset of OFDM subcarriers and the spikes payloads are transmitted via pulse-amplitude modulation (PAM) on all the assigned subcarriers. This way, the addresses are implicitly transmitted via the location of the PAM symbols across the subcarrier indices. While no spikes are dropped as long as the number of subcarriers is large enough, analog transmission may degrade the quality of the reconstructed spikes due to the reliance of repetition coding.
    \item We evaluate the performance of the proposed neuromorphic wireless split computing architecture based on multi-level SNNs both via simulations and via a basic prototype using a neuromorphic camera \cite{hu2021v2e} and USRP boards. 
\end{itemize}

\subsection{Organization}
The remainder of the paper is organized as follows. Section \ref{sec:background} presents  background information about multi-level SNN. Section \ref{sec: system} describes the  neuromorphic wireless split computing system with multi-level spikes under study, while the proposed digital and analog transmission schemes are described in Section \ref{sec: modulation}. Section \ref{sec: receiver} explains neuromorphic receiver processing, including channel estimation, equalization, and decoding SNN processing. Experimental setting and results are described in Section \ref{exp}. Finally, Section \ref{con} concludes the paper.

\section{Multi-Level Spiking Neural Networks} \label{sec:background}
\begin{figure*}[ht!]
    \centering
    \includegraphics[width=0.9\linewidth]{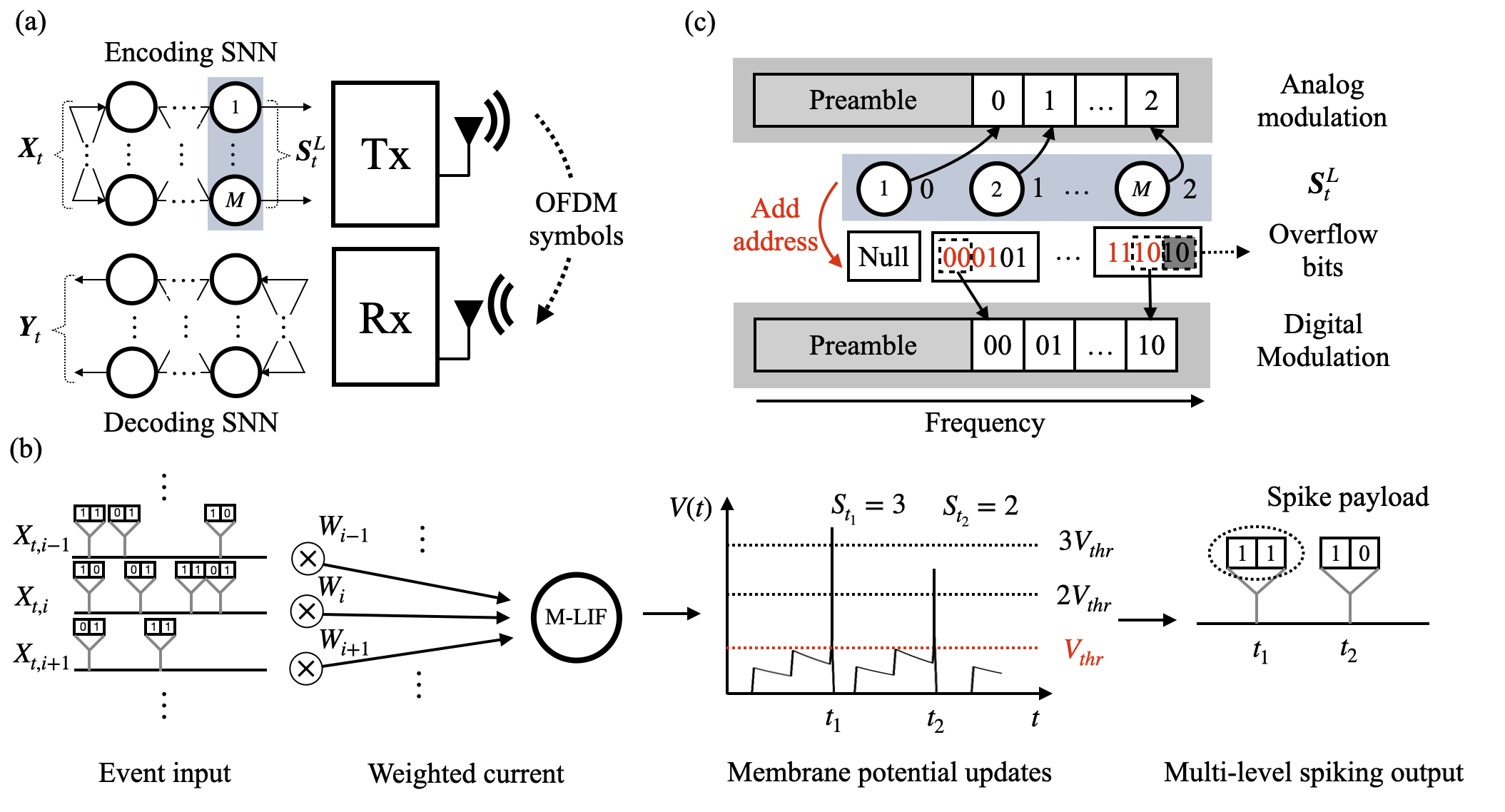}
    \caption{Neuromorphic wireless split computing with multi-level spikes: (a) An SNN is split into an encoding SNN and a decoding SNN, which are connected over a wireless channel following a spilt computing architecture.    (b) Unlike prior works \cite{skatchkovsky2020end, chen2023neuromorphic, chen2022neuromorphic, 10682971}, the SNNs implement spiking neurons that communicate using multi-level spikes \cite{shrestha2024efficient, theilman2024spiking}, adopting a multi-level leaky integrate-and-fire (M-LIF) neuron model. (c) The output of the encoding SNN is transmitted using either analog or digital modulation. In the analog implementation, each output neuron of the encoding SNN is assigned separate OFDM subcarriers. In contrast, in the digital implementation, the AER protocol is used to embed information about the neurons' identities. Overflow bits that do not fit the allocated OFDM symbols are discarded.}
    \label{fig: wireless_ch}
\end{figure*}

As illustrated in Fig. \ref{fig: wireless_ch}, this paper studies a neuromorphic wireless split computing system that leverages spiking neuronal models with multi-level, or graded, spike. In this section, we describe first the conventional SNN model based on leaky integrate-and-fire (LIF) neurons (see, e.g., \cite{10606014, wu2024direct}), and then cover the generalized SNN model with multi-level LIF (M-LIF) neurons \cite{shrestha2024efficient,theilman2024spiking}. The following section will present the proposed split computing architecture based on SNNs  with M-LIF neurons.

\subsection{Conventional Leaky Integrate-And-Fire Neuron}
A conventional LIF neuron accumulates stimuli over time, using an internal state known as membrane potential, and emits a spike once its membrane potential exceeds a certain threshold. LIF neurons can be arranged into arbitrary architectures, and they have been used to implement models such as multi-layer perception transformers \cite{song2024xpikeformer} and  state-space models \cite{shen2024spikingssms}. In this work, we consider an arbitrary layered architecture, in which each neuron $i$ in layer $l$ generates a spike at time $t$ if the local membrane potential $V_{t,i}^l$ passes the threshold $V_{\mathrm{thr}}^l$. Accordingly, the output of the LIF neuron $i$ in layer $l$ at time $t$ is given by 
\begin{align}
    S_{t,i}^l=\left\{
\begin{array}{ll}
0 ~\text{(no spike)},       & \text{if}~ V_{t,i}^l \leq V^l_{\mathrm{thr}}, \\
1 ~\text{(spike)},         & \text{if}~ V_{t,i}^l > V^l_{\mathrm{thr}}.
\end{array} \right. \label{eq: conventional_spike}
\end{align}
The membrane potential $V_{t,i}^l$ is updated via the leaky integrator dynamics
\begin{align}
    V_{t,i}^l= \delta V^l_{t-1,i}(1-S^l_{t-1,i}) + Z^l_{t,i}, \label{eq:membrane}
\end{align}
where $0<\delta<1$ represents the decay factor, and $ Z^l_{t,i}$ denotes the input current from the pre-synaptic neurons in the previous, $(l-1)$-th, layer. By \eqref{eq:membrane}, if the neuron $i$ at layer $l$ spikes at time $t-1$, i.e., if $S_{t-1,i}^l=1$, the membrane potential is reset at time $t$.

Given the vector $\mv S_t^{l-1}$ collecting all the  binary outputs $\{S_{t,i}^{l-1}\}_{i}$ produced in the $(l-1)$-th layer, the input current vector $\mv Z_t^l$, collecting the currents $\{Z_{t,i}^l\}_i$ feeding into each neuron $i$ in layer $l$, is given by the linear combination
\begin{align}
    \boldsymbol{Z}^l_t =   \boldsymbol{W}^l \boldsymbol{S}^{l-1}_t,  \label{syp}
\end{align} 
where $\boldsymbol{W}^l$ is weight matrix between the $(l-1)$-th and the $l$-th layer.  

Evaluating \eqref{syp} requires one accumulate operation per spike, whose energy cost we denote as $E_{\mathrm{ac}}$.

\subsection{Leaky Integrate-and-Fire Neuron with Multi-level Spikes} \label{sec: graded_spike}

Spikes emitted in a conventional LIF-based SNN carry information only via their firing time $t$ defined as in \eqref{eq: conventional_spike}. A more general model, implemented in neuromorphic chips, such as Intel's Loihi \cite{shrestha2024efficient}, allows each spike to carry $m$ additional bits of information. According to Fig. \ref{fig: wireless_ch}, a multi-level, or graded, spike is assigned a payload containing $m$ bits of information.
In this way, when $m=0$, a graded spike reduces to the spike produced by conventional LIF models.  

As reported in \cite{shrestha2024efficient,theilman2024spiking}, digital neuromophic chips such as Intel's Loihi 2 permit spikes to carry integer-valued payloads with marginal additional energy cost.  This is due to several architectural optimizations:
\begin{itemize}
    \item Efficient spike processing: Digital chips commonly use the AER protocol for communications between cores \cite{shrestha2024efficient}. Each AER packet contains address bits, and the few additional bits required to transmit the spike payload -- typically less than 8 bits -- yield a marginal increase the overall communication cost.
    \item Optimized synapse processing: As discussed in \cite{orchard2021efficient}, incoming spikes can be  mapped to lists of synapse weights that are accumulated for consumption in the next time step. This approach allows for the efficient processing of multi-level spikes without significantly increasing computational complexity.
    \item Peripheral modifications: The overhead to generate multi-bit spikes after multiply-accumulate (MAC) operations primarily involves modifying the sense amplifier and ADC \cite{mehonic2024roadmap}. This overhead scales linearly, contrasting with the quadratic scaling of the number of MAC operations being performed.
\end{itemize}

Like LIF neurons, multi-level LIF (M-LIF) neurons produce a spike any time the membrane potential crosses a threshold. However, the payload of the spike produced by an M-LIF neuron contains a payload of $m$ bits, which is obtained by quantizing the membrane potential at the time of spiking.

To elaborate, consider a neuron $i$ in layer $l$ of an arbitrary layered architecture, whose membrane potential is denoted by $V_{t,i}^l$. The output $S_{t,i}^l$ associated with neuron $i$ in layer $l$ at time $t$ is given by  \cite{hao2024lm}
\begin{equation}
S_{t,i}^l=\left\{
\begin{array}{ll}
0~\text{(no spike)},        & \text{if}~ V_{t,i}^l \leq V^l_{\mathrm{thr}}, \\
\mathcal{Q}^l(V^l_{t,i}),       &   \text{if}~ V_{t,i}^l > V^l_{\mathrm{thr}},
\end{array} \right. \label{eq: graded_spike}
\end{equation}
where $\mathcal{Q}(\cdot)$ is an $m$-bit quantizer. By \eqref{eq: graded_spike}, a spike contains $m$ bits given by the quantization level $\mathcal{Q}^l(V^l_{t,i})$. We specifically adopt the clipped uniform quantizer 
\begin{equation}
    \mathcal{Q}^l(V^l_{t,i}) = \min(\lfloor \alpha^l V_{t,i}^l 2^m\rfloor, \\ 2^m) \in \{1,\cdots,2^m\}, \label{eq:uniform_quantizer}
\end{equation}
where $\lfloor \cdot \rfloor$ is the floor operation, and $\alpha^l \in (0,1)$ is a per-layer trainable scaling factor. The integer \eqref{eq:uniform_quantizer}, which ranges in the interval $\{1,\dots, 2^m\}$, is the payload of a spike, which corresponds to $m$ bits.

Evaluating the input currents \eqref{syp} requires the evaluation of a multiply-and-accumulate operation per spike, with the multiplication involving an $m$-bit number. The energy consumption for this operation can be modeled as
\begin{equation}
    E_{\rm mac}(m)=(1+\gamma (m-1))E_{\rm ac}, \label{energy_model}
\end{equation}
where $ \gamma \in [0,1]$ is a technology-dependent parameter. When $\gamma=0$, the hardware optimizations mentioned above are maximally efficient, making $m$-bit synaptic operations as costly as with $m=0$, i.e., $E_{\rm mac}= E_{\rm ac}$. In contrast, when $\gamma = 1$, the energy overhead reflects a basic shift-and-accumulate implementation, which entails an energy cost that is $m$ times that of a single accumulate operation, e.g., $E_{\rm mac}(m)=mE_{\rm ac}$.

\subsection{Training Multi-Level SNN Models} \label{sec: graded_spike_bp}

In this work, we assume the availability of a pre-trained multi-level SNN model, which is split between encoder and decoder as discussed in the next section. Training of the multi-level SNN is achieved using backpropagation via a surrogate gradient that provides a smooth approximation for the hard quantization function in \eqref{eq:uniform_quantizer} \cite{bellec2018long,bengio2013estimating, bu2022optimal}.

To address the non-differentiability of  \eqref{eq: graded_spike}, we first recall the surrogate gradient method used in \cite{bellec2018long}, which applies to a conventional LIF model \eqref{eq: conventional_spike}. This method approximates the derivative of the neuron's output $S^{l}_{i,t}$ in \eqref{eq: conventional_spike} with respect to membrane potential $V_{t,i}^l$  as 
\begin{equation}
\frac{\partial {S}^{l}_{i,t}}{\partial V_{t,i}^l} \approx \Gamma \max\left(0, 1 - \left|V_{t,i}^l - V_{\mathrm{thr}}^l\right|\right), \label{eq:gradient-1}
\end{equation}
where $\Gamma > 0$ is a hyper-parameter. The equation \eqref{eq:gradient-1} replaces the true derivative of the output \eqref{eq: conventional_spike} --- a Dirac delta function at threshold $V_{\mathrm{thr}}^l$ --- with a triangular function centered at the threshold $V_{\mathrm{thr}}^l$ with height $\Gamma$.

\begin{figure}[t!]
    \centering
    \includegraphics[width=0.5\linewidth]{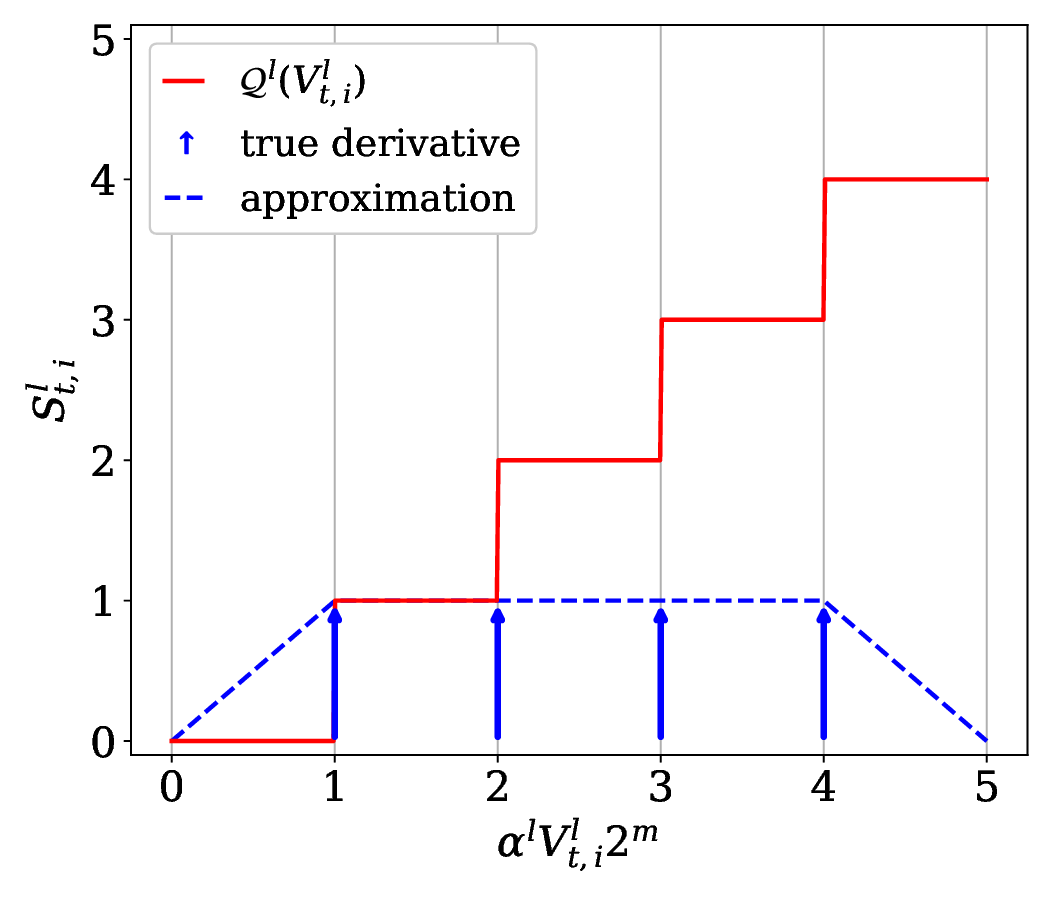}
    \caption{Surrogate derivative \eqref{eq:gradient-2} used for training  SNN models with M-LIF neurons ($m=2$ and $\Gamma=1$).}
    \label{fig:surrogae_gradient}
\end{figure}

To extend this approach to the M-LIF model \eqref{eq: graded_spike}, we first set for simplicity the threshold as $1/\alpha^l2^m$. Then, the derivative of the neuron's output is approximated as shown in Fig. \ref{fig:surrogae_gradient} as
\begin{equation}
\frac{\partial {S}^{l}_{i,t}}{\partial V_{t,i}^l}  \approx \hspace{-0.2em}
\begin{cases} 
      \Gamma \max\left(0, 1 - \left|\alpha^l V_{t,i}^l2^m - 1 \right|\right), & \hspace{-1em} \text{if }  V_{t,i}^l < \frac{1}{\alpha^l2^m}, \\[10pt]
      1, & \hspace{-1em} \text{if } \frac{1}{\alpha^l 2^m} \leq  V_{t,i}^l \leq \frac{1}{\alpha^l}, \\[10pt]
      \Gamma \max\left(0, 1 - \left|\alpha^l V_{t,i}^l 2^m - 2^m\right|\right), & \hspace{-0.8em} \text{if } V_{t,i}^l > \frac{1}{\alpha^l} - \frac{1}{\alpha^l 2^m}.
\end{cases}\label{eq:gradient-2}
\end{equation} 
As illustrated in Fig. \ref{fig:surrogae_gradient}, the discontinuous, impulsive, derivatives, associated with the multi-level spikes \eqref{eq: graded_spike} $S_{t,i}^l$ are approximated in a manner that extends (\ref{eq:gradient-1}) via a piece-wise function \cite{bengio2013estimating, bu2022optimal}.

\section{Neuromorphic Wireless Split Computing With Multi-Level Spikes} \label{sec: system}
In this section, we describe the neuromorphic split computing system under study. In order to accommodate a multi-bit spike payload, as well as to facilitate prototyping using SDR technology, we adopt an OFDM radio interface in lieu of the UWB modulation considered in prior works \cite{10682971, skatchkovsky2020end, chen2023neuromorphic, chen2022neuromorphic}.

\subsection{Neuromorphic Wireless Split Computing}
As illustrated in Fig. \ref{fig: wireless_ch}(a), we consider a neuromorphic wireless split computing system in which an SNN consisting of M-LIF neurons is split between a single-antenna transmitter (Tx) and a single-antenna receiver (Rx). The SNN is trained to solve an inference problem based on data captured by a neuromorphic sensor, such as a dynamic vision sensor (DVS) camera \cite{hu2021v2e} at the Tx. Based on the SNN split, the Rx makes the final inference decision by using the wireless signals received from the Tx. Practical examples of applications of this architecture were presented in \cite{lee2024asynchronous,he2022implantable}, including brain-computer interfaces and biomedical devices. 

Neuromorphic sensors generate a spike whenever a relevant event is detected, such as a significant change in pixel brightness. Spikes produced by the sensor are often graded, e.g., indicating the sign of the reported change with a one-bit payload \cite{hu2021v2e}.  

\begin{figure}[ht!]
    \centering
    \includegraphics[width=0.75\linewidth]{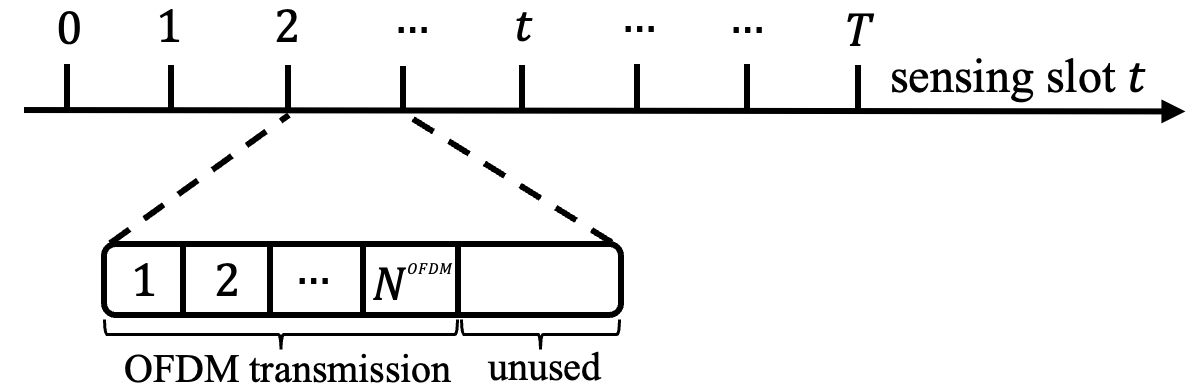}
    \caption{Timeline of the proposed neuromorphic wireless split computing system. Time is discretized into sensing slots $t=1,2,\ldots,T$, corresponding to the time period over which the neuromorphic sensor accumulates information before reporting the presence or absence of events, along with the corresponding payloads. The spikes produced at time slot $t-1$ are processed by the Tx, and the outputs of the encoding SNNs are transmitted over the air using $N^{\rm OFDM}$ OFDM symbols to the Rx during the following, $t$-th, sensing slot. The decoding SNN at the Rx then processes the received signals to produce an inference decision. Each sensing time  step $t$ is typically much longer than the duration of an OFDM symbol.}
    \label{fig: timeline}
\end{figure}

As illustrated in Fig.~\ref{fig: timeline}, we adopt a discrete-time model, where time is divided into sensing slots indexed by integers $t=1,2,\ldots$ Each slot corresponds to the time period over which the neuromorphic sensor accumulates information before reporting the presence or absence of events, along with the corresponding payloads. 

Accordingly, at the end of each sensing period $t$, the neuromorphic sensor at the Tx produces a $D \times 1$ vector $\mv X_t=[X_{t,1}, \ldots, X_{t, D}]^T$ representing multi-level spikes produced by each of the $D$ elements of the sensor. For example, a DVS camera produces $D$ signals, each corresponding to one pixel. Each entry $X_{t,i}$ represents the presence $(X_{t,i}>0)$ or absence $(X_{t,i}=0)$ of a spike. When a spike is present, the payload $X_{t,i}\in\{1,\ldots,2^m\}$ encompasses $m$ bits.  

The spiking signal $\mv X_t$ recorded by the neuromorphic sensor is processed by the encoding SNN. Unlike prior works \cite{10682971, skatchkovsky2020end, chen2023neuromorphic, chen2022neuromorphic}, the encoding and decoding SNNs in the proposed system are capable of processing and producing multi-level spikes with $m$ bits. Specifically, as introduced in the previous section, we consider arbitrary layered architectures consisting of M-LIF neurons for both the encoding and decoding SNNs. 

Denote by $M$ the number of neurons in the last layer of the encoding SNN, and by $ \boldsymbol{S}_t = [S_{t,1}, S_{t,2}, \dots, S_{t,M}]$ the vector of graded spikes produced by the output layer of the encoding SNN, with $S_{t,i}\in \{0,1,\ldots,2^m\}$ for all $i\in\{1,\ldots,M\}$. The multi-level spikes $\boldsymbol{S}_t$ generated by the encoding SNN are modulated by the transmitter onto a baseband signal that is transmitted over a wireless channel using OFDM. Using the received signals, the decoding SNN at the Rx produces the final inference decision.

\subsection{OFDM Transmission of Multi-Level Spikes} \label{sec: ofdm}
 
As illustrated in Fig.~\ref{fig: timeline}, a number $N^{\rm OFDM}$ of OFDM symbols is available in each sensing slot $t$ to transmit information about the previous slot $t-1$. The duration of each sensing slot is typically sufficient to accommodate a large number of OFDM symbols, and we assume that the wireless interface is used for other devices and/or services when not occupied by the Tx. For instance, a DVS camera has a typical sensing period of $130$ ms \cite{amir2017low}, while an OFDM symbol for a 5G wireless link with a bandwidth 20 MHz takes 38.09 $\mu s$.

Each OFDM symbol consists of $N^{\rm D}+N^{\rm P}$ subcarriers, with $N^{\rm D}$ data subcarriers and $N^{\rm P}$ pilot subcarriers. We denote the subset of data subcarriers as $\mathcal{N}^{\rm D}$ and the subset of pilot subcarriers as $\mathcal{N}^{\rm P}$, with $|\mathcal{N}^{\rm D}|=N^{\rm D}$ and $|\mathcal{N}^{\rm P}|=N^{\rm P}$. 

Assuming that the cyclic prefix (CP) of each OFDM symbol is no shorter than the discrete delay spread of the multi-path channel, the $n$-th received OFDM symbol at sensing time period $t$ is given by \cite{goldsmith2005wireless}
\begin{align}
    \mv y^n_t= \mv H^n_t \mv x^n_t + \mv w^n_t, \label{receiving}
\end{align}
where the diagonal channel matrix $\mv H^n_t$ collects on its main diagonal the channel frequency responses across all the subcarriers; $\mv x_t^n$ is the $n$-th frequency-domain OFDM symbol encompassing both the pilot signals and the payload of the  multi-level spikes for slot $t-1$; and $\mv w^n_t$ is a noise vector, with independent and identically distributed (i.i.d.) complex Gaussian elements having zero means and variances $N_0$.

\section{Digital and Analog Transmission of Multi-Level Spikes} \label{sec: modulation}
In this section, we present digital and analog modulation strategies for encoding multi-level spikes at each sensing time slot. We begin by introducing pilot transmission, which enables channel estimation for OFDM symbol equalization. Next, we discuss two different types of power constraints, and, finally, we detail the digital and analog modulation schemes.

\subsection{Pilot Transmission}
To enable the receiver to perform effective channel estimation and equalization for both analog and digital modulation, the Tx sends pilot symbols $\{x_{t,i}^{\rm p}\}_{i \in \mathcal{N}^{\rm P}}$, known to the Rx, at a given power level $P^{\rm max}$ on designated pilot subcarriers indexed by the integers $i\in \mathcal{N}^{\rm P}$. As discussed in the next section, the pilot subcarriers serve as references to obtain a channel estimate, which is leveraged for the equalization of data symbols.

\subsection{Power Constraints}
At each sensing slot $t$, the graded spikes $\mv S_{t-1}$ are modulated into symbols $\{x_{t,i}^n\}_{i\in\mathcal{N}^{\rm D}}$ transmitted on the subset $\mathcal{N}^{\rm D}$ of data subcarriers in each OFDM symbol indexed as  $n=1,2,\ldots, N^{\rm OFDM}$.
We consider two types of power constraints on the data subcarriers.
\subsubsection{Average Per-Symbol Power Constraint} The average per-symbol power constraint limits the overall power used for transmission across all data subcarriers in a given OFDM symbol. Denoting the collection of $N^{\rm D}$ symbols transmitted on all data subcarriers of the $n$-th OFDM symbol at sensing slot $t$ by $\mv x^n_{t}$, this constraint is given by
\begin{equation}
   \frac{1}{N^{\rm D}} \|\mv x^n_{t}\|^2 \leq  P^{\rm max},  \label{eq:energy}
\end{equation}
where $P^{\rm max}$ represents the maximum allowable average transmission power per data subcarrier. This constraint ensures that the total transmitted power remains within acceptable limits, helping to control energy consumption and interference. 

\subsubsection{Peak Power Constraint} While the average per-symbol power constraint \eqref{eq:energy} allows for adaptive power allocation among data subcarriers, in practice, the Tx has a maximum power limit that cannot be exceeded for each subcarrier. To address this type of constraint, we also study a peak power requirement, which restricts the maximum power transmitted on each data subcarrier to a fixed value $P^{\rm max}$, i.e.,
\begin{equation}
    \|x^n_{t,i}\|^2 \leq P^{\rm max}  \label{eq:energy2}
\end{equation}
for each subcarrier $i$.
This peak power constraint helps maintain the peak-to-average power ratio (PAPR) within reasonable bounds, thereby mitigating potential issues like amplifier non-linearities, which can adversely affect signal quality and system performance \cite{goldsmith2005wireless}.

\subsubsection{Dynamic Power Constraints} Under either constraint \eqref{eq:energy} or \eqref{eq:energy2}, the total maximum power allocated at each slot $t$ remains the same, i.e.,
\begin{equation}
   \frac{1}{N^{\rm D}N^{\rm OFDM}} \sum_{n=1}^{N^{\rm OFDM}} \|\mv x^n_{t}\|^2 \leq  P^{\rm max}. 
\end{equation}
Accounting for the sequential nature of processing in SNNs, we also explore dynamic power allocation strategy across different sensing slots. Specifically, denoting as $P_t$ the power allocated in slot $t$, we impose the total power budget constraint
\begin{align}
    \frac{1}{T} \sum_{t=1}^T P_t \leq P^{\rm max}, \label{limit}
\end{align}
across $T$ time slots. This enables the dynamic allocation of transmit power levels $P_t$ over the time index $t$. Within each slot $t$, the power constraints \eqref{eq:energy} or \eqref{eq:energy2} is enforced with the power $P_t$ in lieu of $P^{\rm max}$.

\subsection{Digital Modulation of Multi-Level Spikes}\label{sec: digital_modulation}
For digital modulation, we adopt a standard AER protocol to encode the multi-level spikes \cite{perez2010fast}. Accordingly, each graded spike $S_{t,i}>0$ at time $t$ is associated with an AER packet containing the address of the $i$-th spiking output neuron, as well as with the payload of $m$ bits. The address of the $i$-th output neuron is encoded into  $\lceil\log_2(M)\rceil$ bits, as there are $M$ possible spiking neurons at the output layer. Hence, The resulting AER packet contains $\lceil\log_2(M)\rceil+m$ bits. 

Following the timeline in Fig.~\ref{fig: timeline}, the AER packets corresponding to all spikes generated at each sensing slot $t-1$ from the output layer of the encoding SNN are encoded and modulated using $N^{\rm OFDM}$ OFDM symbols at sensing slot $t$. The $N^{\rm OFDM}$ symbols must thus ideally encode a number of AER packets equal to $\sum_{i=1}^M \mathbbm{1}(S_{t-1,i}>0)$. This yields a total of 
\begin{equation}
    B^{\rm ToT}_t=(\lceil\log_2(M)\rceil+m)\sum_{i=1}^M\mathbbm{1}( S_{t-1,i} >0)
     \label{eq:totbit}
\end{equation}
bits to be transmitted. However, due to the varying level of sparsity of the output neurons at different times $t-1$, the allocated OFDM symbols may be insufficient to deliver all these bits. 
In particular, using a binary channel code with rate $0<r \leq 1$ and a modulation scheme with $2^B$ constellation points, the number of bits that can be transmitted by the Tx to the Rx is $B^{\rm OFDM}=N^{\rm OFDM}N^{\rm D}Br$.

If the spiking signals of the output neurons are sparse enough, so that the number of bits $B^{\rm ToT}_t$ does not exceed the capacity $B^{\rm OFDM}$ of the available OFDM symbols, all the AER packets are encoded for transmission. Otherwise, the largest subset $\mathcal{S}_t$ of AER packets is selected so as to guarantee the condition $(\lceil\log_2(M)\rceil+m)|\mathcal{S}_t| \leq B^{\rm OFDM}$. The subset  $\mathcal{S}_t$ is selected uniformly at random among the $\sum_{i=1}^M \mathbbm{1}(S_{t-1,i}>0)$ AER packets. Note that, in case the output signals are very sparse, i.e., if $B^{\rm ToT}_t < B^{\rm OFDM}$, some subcarriers remain unused.

\subsection{Analog Modulation of Multi-Level Spikes} \label{analogm}
As discussed in the previous subsection, digital modulation represents each multi-level spike using an AER format, requiring additional bits to specify addresses, as well as redundant bits for channel coding. In contrast, as detailed next, analog modulation directly maps the multi-level spikes onto a set of PAM symbols. These PAM symbols are then loaded onto the OFDM symbols by mapping subcarriers to output neurons of the encoding SNN, without the need for addressing or coding overhead.

With analog modulation, each of the $M$ output neurons of the encoding SNN is assigned to a subset of data subcarriers in the OFDM symbols corresponding to the current time slot $t$. We denote the mapping between neuron $i$ and a subset of subcarriers $\mathcal{N}^{\rm D}(i) \subseteq \mathcal{N}^{\rm D}$. The subsets $\mathcal{N}^{\rm D}(i)$ are disjoint, so that each subcarrier is uniquely assigned to one neuron. This requires the assumption $N^{\rm D}N^{\rm OFDM} \geq M$. Moreover, each set $\mathcal{N}^{\rm D}(i)$ contains $\lceil N^{\rm D}N^{\rm OFDM}/M \rceil$ subcarriers.  

Each multi-level spike $S_{t,i}>0$ from output neuron $i$ at sensing slot $t$ is mapped into a $2^m$-PAM symbol, which is transmitted on all subcarriers in subset $\mathcal{N}^{\rm D}(i)$. The transmission of the same symbol on multiple subcarriers amounts to a form of repetition coding. In contrast, if there is no spike, i.e., if $S_{t,i}=0$, the corresponding subcarriers in subset $\mathcal{N}^{\rm D}(i)$ are idle. Therefore, sparser spiking signals entail a larger number of unused subcarriers.

\section{Neuromorphic Receiver Processing} \label{sec: receiver}
In this section, we describe the processing applied by the receiver for both digital and analog modulation schemes. The receiver first estimates the channel using standard signal processing methods, allowing for equalization of the received symbols. Once equalization is complete, the receiver demodulates the data symbols to estimated multi-level spikes, which are then passed to the decoding SNN to make the final inference decision.

\subsection{Channel Estimation} \label{sec:ce}
Denote by $\mv h^n_t=[H^n_{t,1}, \ldots, H^n_{t,N+N^{\rm p}}]^T$ the diagonal elements of the frequency-domain channel matrix $\mv H^n_t$ in \eqref{receiving}. Note that $H^n_{t,i}$ corresponds to the channel gain for the $i$-th subcarrier in the $n$-th OFDM symbol at time $t$. Following the standard approach \cite{goldsmith2005wireless}, the receiver first estimates the channel vector $\hat{\mv h}^n_t$ on the pilot subcarriers. Based on these estimates, the channel gains for the data subcarriers are determined via interpolation \cite{1259423}.

Specifically, the receiver extracts from the received signal $\mv y^n_t$ in \eqref{receiving} the received pilot symbols $\mv y^{\rm p}_{t,n}=\{y^n_{t,i}\}_{i\in\mathcal{N}^{\rm P}}$. The channels on the pilot subcarriers are estimated using the least squares (LS) method. The resulting estimated channel $\hat{h}^n_{t,i}$ for pilot subcarrier $i\in\mathcal{N}^{\rm P}$ of the $n$-th OFDM symbol is given by 
\begin{align}
    \hat{h}^n_{t,i}=\frac{y^n_{t,i}}{x_{t,i}^{\rm p}},\label{eq:LS}
\end{align}
where $x_{t,i}^{\rm p}$ is the known pilot symbol transmitted at time $t$ on pilot subcarrier $i\in\mathcal{N}^{\rm P}$.

With the estimated channels $\{\hat{h}^n_{t,i}\}_{i\in\mathcal{N}^{\rm P}}$ on the pilot subcarriers for each $n$-th OFDM symbol, the channel response on the data subcarriers is estimated via linear interpolation. To elaborate, write the set of pilot subcarriers $\mathcal{N}^{\rm P}=\{i_1,i_2,\ldots,i_P\}\subset\{1,2,\ldots,N+N^{\rm p}\}$ and assume pilots are interleaved with data subcarriers. For a data subcarrier $i\in\mathcal{N}^{\rm D}$, we find the largest pilot subcarrier $i_p<i$ and the smallest pilot subcarrier $i_{p+1}>i$, where $i_p<i<i_{p+1}$. The estimated channel response for the data subcarrier $i\in\mathcal{N}^{\rm D}$ at time $t$ is then given by
\begin{align}
    \hat{h}^n_{t,i} = \hat{h}^n_{t,i_p} + \frac{\hat{h}^n_{t,i_{p+1}}-\hat{h}^n_{t,i_p}}{i_{p+1}-i_p}\big(i-i_p\big).
\end{align}

\subsection{Equalization}
After obtaining the channel estimate, zero forcing (ZF) equalization is applied to mitigate the effect of the channel on the received symbols. Specifically, the equalized data symbol for the $i$-th data subcarrier in the $n$-th OFDM symbol at sensing slot $t$ is given by the ratio
\begin{align}
    \hat{x}^n_{t,i} = \frac{y^n_{t,i}}{\hat{h}^n_{t,i}}. \label{eq:equalization}
\end{align}

For both digital and analog modulation, each equalized symbol $\hat{x}^n_{t,i}$ is demodulated into one of the $2^B$ constellation points using maximum likelihood detection. This detection involves selecting the constellation point that is closest to the equalized symbol in terms of Euclidean distance, ensuring that the received symbol is mapped to its most likely transmitted value.

\subsubsection{Digital Implementation} \label{dim}
For the digital implementation, each equalized symbol $\hat{x}^n_{t,i}$ is demodulated into one of the $2^B$ constellation points, resulting in a $B$-bit sequence. The bits from all data subcarriers over $N^{\rm OFDM}$ OFDM symbols form a total of $N^{\rm OFDM}N^DB$ bits. These bits are then processed through channel decoding to reconstruct the estimated AER packets.

Each estimated AER packet contains a payload of $m$ bits and an address $i$, which are mapped to the estimated multi-level spike $\hat{S}_{t,i}$.  If no multi-level spike is detected for a neuron $i$, the neuron's input is set to $\hat{S}_{t,i}=0$, indicating no activity. This results in a vector $\hat{\mv S}_t=[\hat{S}_{t,1}, \hat{S}_{t,2}, \ldots, \hat{S}_{t,M}]^T$ of estimated spikes, which serves as the input to the decoding SNN.

\subsubsection{Analog Implementation}In the analog implementation, the $\lceil N^{\rm D}N^{\rm OFDM}/M \rceil$ equalized subcarriers corresponding to the same symbol are averaged to enhance signal quality. After averaging, the symbols are demodulated by using maximum likelihood detection to determine the most likely value from a $2^B$-level PAM constellation. The demodulated PAM symbols are then mapped to the estimated multi-level spikes, resulting in the vector $\hat{\mv S}_t=[\hat{S}_{t,1}, \hat{S}_{t,2}, \ldots, \hat{S}_{t,M}]^T$ of estimated spike activities of the $M$ input neurons for the decoding SNN.

\subsection{Decoding SNN Processing}
The estimated spikes $\hat{\mv S}_t=[\hat{S}_{t,1}, \hat{S}_{t,2}, \ldots, \hat{S}_{t,M}]^T$, derived from either analog or digital modulations, are processed by the decoding SNN. For a classification task, the decoding SNN has $C$ output neurons, with each output neuron representing a specific class.

Focusing on classification, this work uses a membrane potential-based decision rule for classification \cite{deng2022temporal}. Following this approach, the classification decision is determined by identifying the output neuron that has the highest integrated membrane potential over all sensing slots $t=1,\ldots,T$.  The decision rule can be expressed mathematically as
\begin{align} 
\hat{c} = \argmax_{i \in \{1, \dots, C\}} \sum_{t=1}^T V_{t,i} , 
\end{align}
where $V_{t,i}$ represents the membrane potential of the $i$-th output neuron at sensing slot $t$.

\section{Simulation Results and Experiments} \label{exp}

In this section, we report results from simulations and real-world experiments with the main aim of investigating the potential advantages of multi-level spikes in neuromorphic wireless split computing\footnote{Code is available at https://github.com/kclip/neurocomm-msnn}.

\subsection{Setting}
\subsubsection{Inference Task}
The system is configured to classify event-based inputs, captured by a neuromorphic vision sensor. Specifically, we consider the standard DVS128 Getsure dataset \cite{amir2017low}, which consists of data logged by a DVS camera for a duration of 6 seconds when presented one out of 11 possible gestures  \cite{4444573}. An SNN with M-LIF neurons having five convolutional layers followed by four fully connected layers is pre-trained by using the approach discussed in Sec. \ref{sec: graded_spike_bp}.
We split the network, configured as 2C4-64C3-128C3-128C3-128C3-128C3-AP2-FC512-FC256-FC128-FC11, at the first fully connected (FC) layer. In this notation, C denotes a convolutional layer, and  AP is the average-pooling layer, with the numbers before and after each layer description specifying the number of input channels and kernel size, respectively. The first six layers form the encoding SNN, while the remaining layers constitute the decoding SNN.
Accordingly, the encoding SNN's output layer contains $M=512$ neurons. We train different SNN models for different pairs $(T,m)$ consisting of number $T\in \{2, 4, 6, 8, 10\}$ of sensing slots and payload size $m\in\{0,2,4,6,8\}$. 

To define the sensing slots, each original recording from the DVS camera is divided into four segments of 1.3 seconds each.
Each segment is further divided into 10 frames, each of duration 130 ms, with each frame representing a single sensing slot. The events within each sensing slot are accumulated and directly fed to the first layer of the SNN. For each sensing slot, each neuron in the hidden layer can emit at most one multi-level spike.

\subsubsection{Simulation Setting} In the simulation results, the number of OFDM symbols per sensing slot is set to $N^{\rm OFDM}=5$, and each OFDM symbol consists of $N^{\rm D}=512$ data subcarriers and $N^{\rm P}=75$ pilot subcarriers. Pilot symbols are interleaved with data subcarriers, so that a pilot is placed every 8 data symbols.
We consider a five-path frequency-selective channel, where each path amplitude follows a Rayleigh distribution with the same average power so that the average channel norm equals 1. The signal-to-noise ratio (SNR) is defined as the ratio of the peak or average per-subcarrier symbol power $P^{\rm max}$ over the noise power, i.e.,  $\text{SNR}=P^{\rm max}/N_0$. If not stated otherwise, the average SNR is set to 25 dB. 

We also consider dynamic power allocation strategies with a exponentially decreasing power
\begin{align}
    P_t= a \cdot b^{T-t}, \label{dypo}
\end{align}
where $b$ determines the exponential decay rate of allocated power, and $a$ is a scaling factor selected to ensure the power constraint \eqref{limit}. The rationale for considering this type of power allocation is that errors made in earlier sensing slots may have a cascading effect, causing further degradation compared to errors affecting later slots. 

For digital transmission, we use quadrature phase shift keying (QPSK) modulation, providing $B=2$ bits per subcarrier, along with low-density parity-check (LDPC) coding with rate $r=1/2$ using the implementation in Nvidia's Sionna \cite{sionna}. After channel estimation, equalization and demodulation, the bit sequence is decoded using the belief propagation algorithm provided in \cite{sionna}. As detailed in Sec. \ref{dim}, the recovered bit stream is mapped to the AER packets, and the corresponding multi-level spikes are fed to the decoding SNN.

For analog transmission, as presented in Sec. \ref{analogm},  each $m$-bit multi-level spike is quantized into one of the $2^m$ PAM constellation points, with the absence of a spike for a neuron corresponding to idle subcarriers. Specifically, each output neuron of the encoding SNN is mapped to one subcarrier in each OFDM symbol. Thus, each PAM symbol is transmitted $N^{\rm OFDM}$ times. The received OFDM symbols are equalized and averaged before being demodulated into PAM symbols. Finally, the demodulated PAM symbols are mapped back to multi-level spikes.

\subsubsection{Experimental Setting}
\begin{figure}[!ht]
    \centering
    \includegraphics[width=0.5\linewidth]{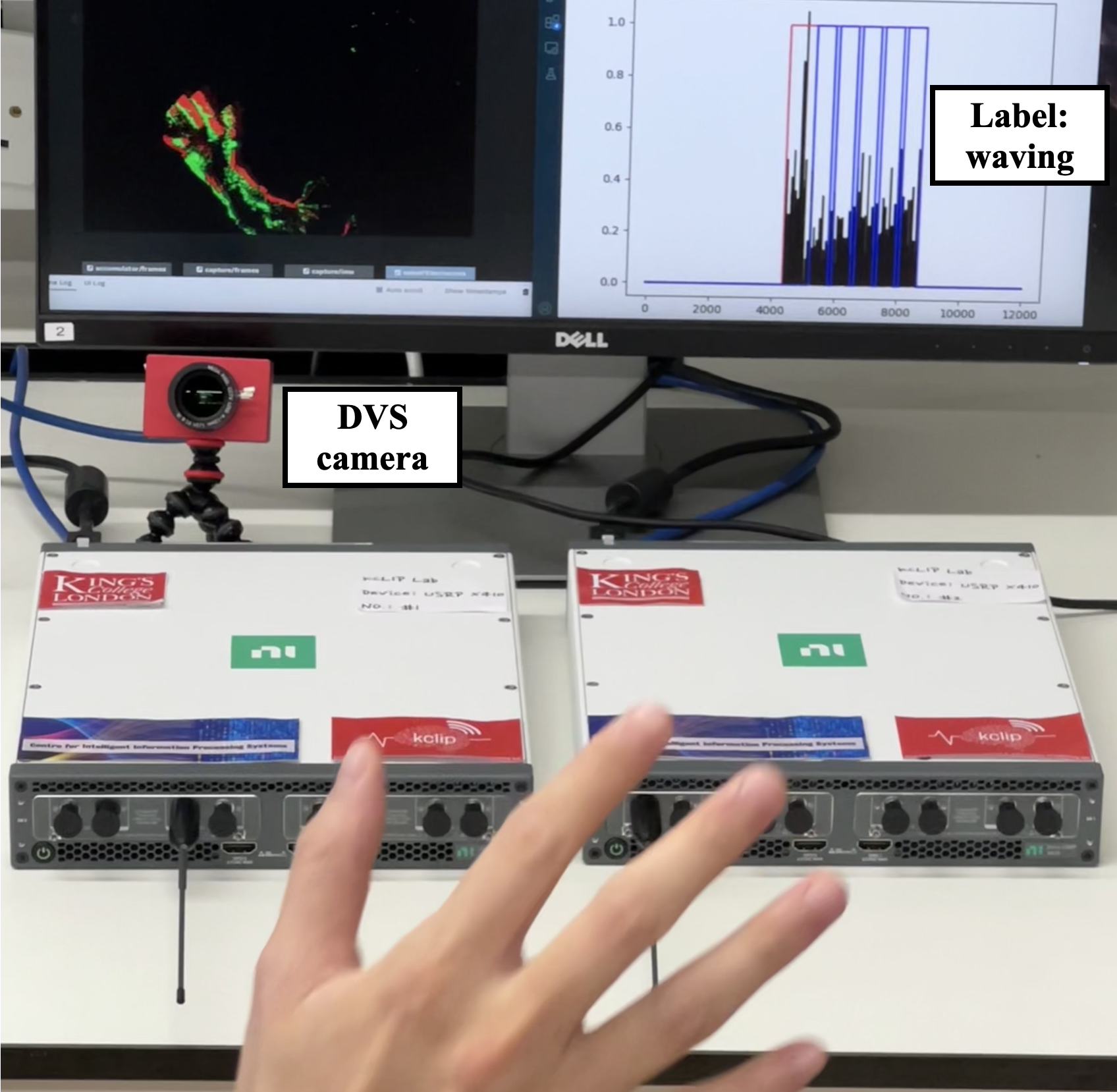}
    \caption{The experimental setup includes a DVS sensor, a transmitter and a receiver. The screen visualizes the event-based input of the DVS sensor (left), along with the corresponding received OFDM signal at the receiver and the gesture type detected by the decoding SNN (right).}\label{fig:usrp_setup}
\end{figure}

To validate the proposed system in a real-world scenario, we implement the proposed end-to-end neuromorphic wireless remote inference system using USRP SDRs \cite{USRP}. As illustrated in Fig. \ref{fig:usrp_setup}, one USRP board is configured as the Tx connected to the DVS camera, while the other USRP board serves as the Rx.

We set up the radios to exchange one frame per sensing slot of $130$ ms. To permit synchronization of the receiver, we assume the transmission of digital pilot signals by following the standard implementation detailed in \cite{syn}. Each frame consists of a known preamble for synchronization, followed by $N^{\rm OFDM}$ OFDM symbols with $N^{\rm D}=512$ data subcarriers and $N^{\rm P}=75$ pilot subcarriers as for the simulations described above. The carrier frequency is set to 3.58 GHz, with the gain configured to $50$ dB, and both the transmit and receive rates set to $10^6$ symbols per second. With these choices, an OFDM symbol lasts for $0.714$ ms, which is significantly shorter than the duration of a single sensing slot. 

The distance between the transmitter and receiver was set to approximately 1 meter in an indoor environment with a line-of-sight (LoS) path, ensuring minimal external interference. The system bandwidth is determined by the sampling rate of $1$ MHz, with the active subcarriers spanning an effective bandwidth of approximately $0.7$ MHz. The signal-to-noise ratio (SNR) at the receiver was observed to be approximately $30$ dB. Note that since the USRP is an uncalibrated device, the configured gain value does not correspond to an exact transmit power level \cite{USRP}.

As mentioned, we select the pre-trained SNN models fixed as a function of the parameters $T$ and $m$ throughout this section. However, as we will detail, we found that this approach does not work well with analog modulation when implemented using SDRs. Therefore,  for analog transmission, we also considered fine-tuning the encoding and decoding SNNs to the given deployment via end-to-end training with measured channels. To this end, we first measure a set of $10,000$ channel samples via the receiving USRP board. These samples are then used to simulate channel transmission during fine-tuning. In order to differentiate the quantization error loss through the PAM modulation mapping producing the transmitted symbols, we approximate the underlying quantizer via a temperature-scaled softmax function as in \cite{10285894}.

\subsection{Results}
\subsubsection{Noiseless Channel}
\begin{figure}[t!]
    \centering
    \includegraphics[width=0.6\linewidth]{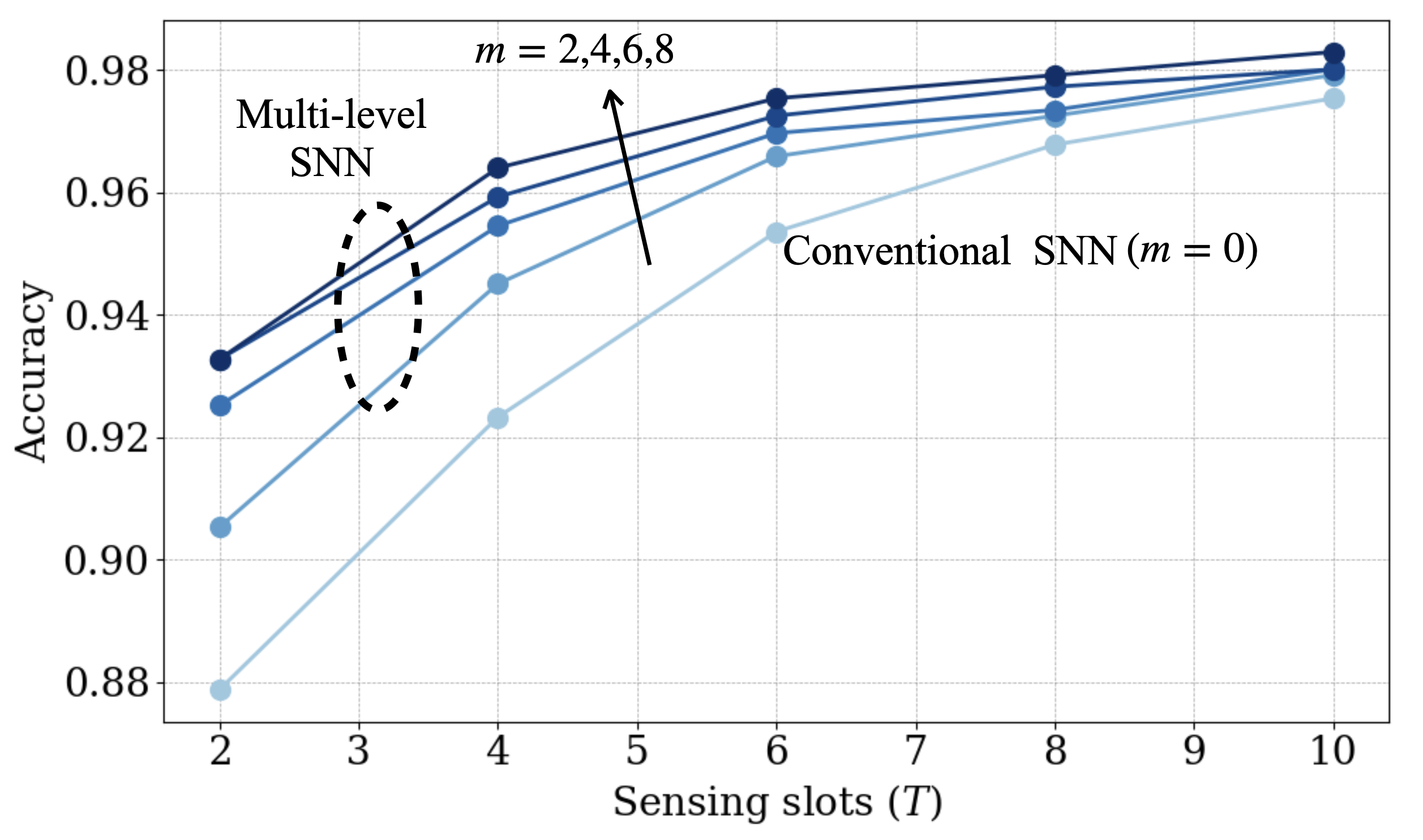}
    \caption{Accuracy versus number of sensing slots $T$ for a centralized implementation of a conventional SNN classifier with $m=0$-bit payloads, as well as for multi-level SNN classifiers with $m=2,4,6,8$-bit payloads.}
    \label{fig:acc_snr_noiseless}
\end{figure}

To start, Fig. \ref{fig:acc_snr_noiseless} illustrates the accuracy performance of the pre-trained SNN model as a function of the number $T$ of sensing slots in a fully centralized implementation. This performance serves as a benchmark for the wireless split computing system to be investigated next. Increasing $T$ enhances the informativeness of the input by extending the sensing period. Accordingly, the accuracy of the SNN classifier increases with the input duration $T$.

The figure compares results obtained with SNN models processing multi-level spikes with a different size $m$ of the spike payload in bits.  Compared to the conventional SNN model with $m=0$, multi-level spikes consistently achieve higher accuracy, particularly at earlier sensing slots. For example, at $T=4$, the conventional SNN model reaches an accuracy of 92.33\%, while a multi-level SNN with $m=2$ attains 94.70\% and $m=8$ yields an accuracy of 96.40\%. 
In contrast, for $T=10$, the conventional SNN achieves an accuracy of 97.54\%, which is similar to the result obtained in \cite{fang2021incorporating} using parametric LIF neurons, while a multi-level SNN achieves 98.30\%.

The outlined performance gains attained with multi-level spikes may not be retained in a wireless split architecture due to an inherent trade-off between the number of transmitted bits and the accuracy of the decoded bits. To illustrate the potential benefits of multi-level spikes in this context, based on Fig.~\ref{fig:acc_snr_noiseless}, in the following we set $T=4$, adopting the corresponding pre-trained SNNs models for different values of $m$. 

To assess inference energy consumption, following \cite{yang2022lead}, we count the number of accumulate operations carried out by the SNN, and adopt the model \eqref{energy_model} described in Section II-B with $E_{\rm ac}=0.1$ pJ \cite{horowitz20141}. For comparison, in a manner similar to \cite{chen2023neuromorphic}, we also consider the performance of an ANN with the same architecture of the SNN with a ReLU activation function and with a softmax output layer. The ANN is trained on the same data. The ANN takes the same input of the SNN at each sensing slot $t$, and makes a final decision by selecting the index with the highest average output of the softmax output layer over $T$ slots. The energy for multiply-and-accumulate operations for the ANNs is set to $E_{\rm mac}=3.2$ pJ \cite{horowitz20141}. 

\begin{figure}[t!]
    \centering
    \includegraphics[width=0.55\linewidth]{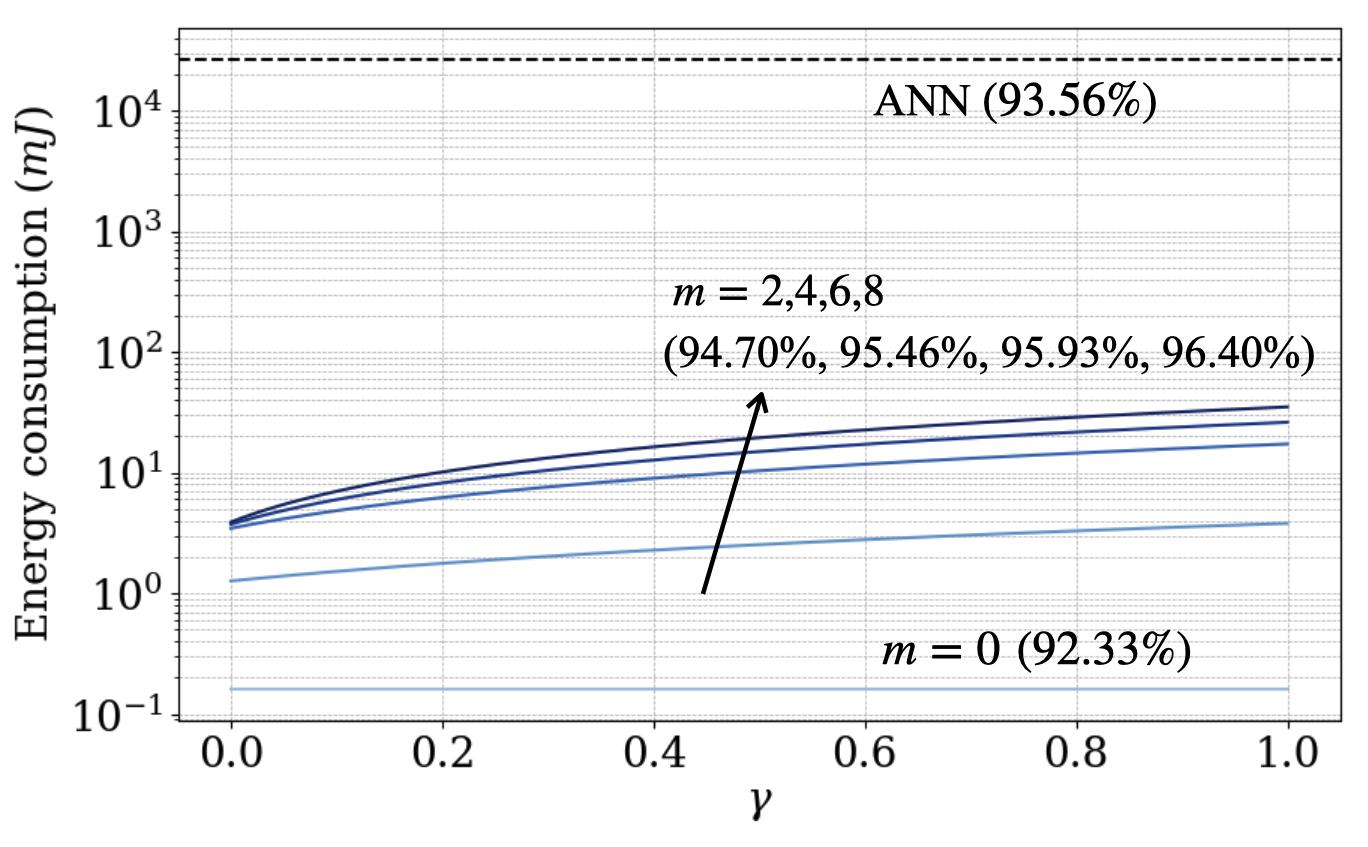}
    \caption{Energy consumption for a conventional SNN ($m = 0$) and for multi-level SNNs ($m = 2,4,6,8$) compared to an ANN with the same architecture. The parameter $\gamma$ reflects the hardware efficiency in processing multi-level spikes, with $\gamma=0$ corresponding to a maximally efficient system. The test accuracy is reported in parenthesis.}
    \label{fig:acc_energy}
\end{figure}

Fig.~\ref{fig:acc_energy} illustrates that increasing the number of bits $m$ from 0 to 8 leads to a modest increase in energy consumption, while yielding a substantial accuracy increase from 92.33\% to 96.40\%. In all cases, the energy consumption remains lower than that of the ANN model. Furthermore, the lower accuracy of the ANN model is due to its lack of a memory mechanism for temporal processing.

\subsubsection{Simulation Results}
\begin{figure*}[t!]
    \centering
    \subfloat[Per-block power constraint]{\includegraphics[width=0.5\columnwidth]{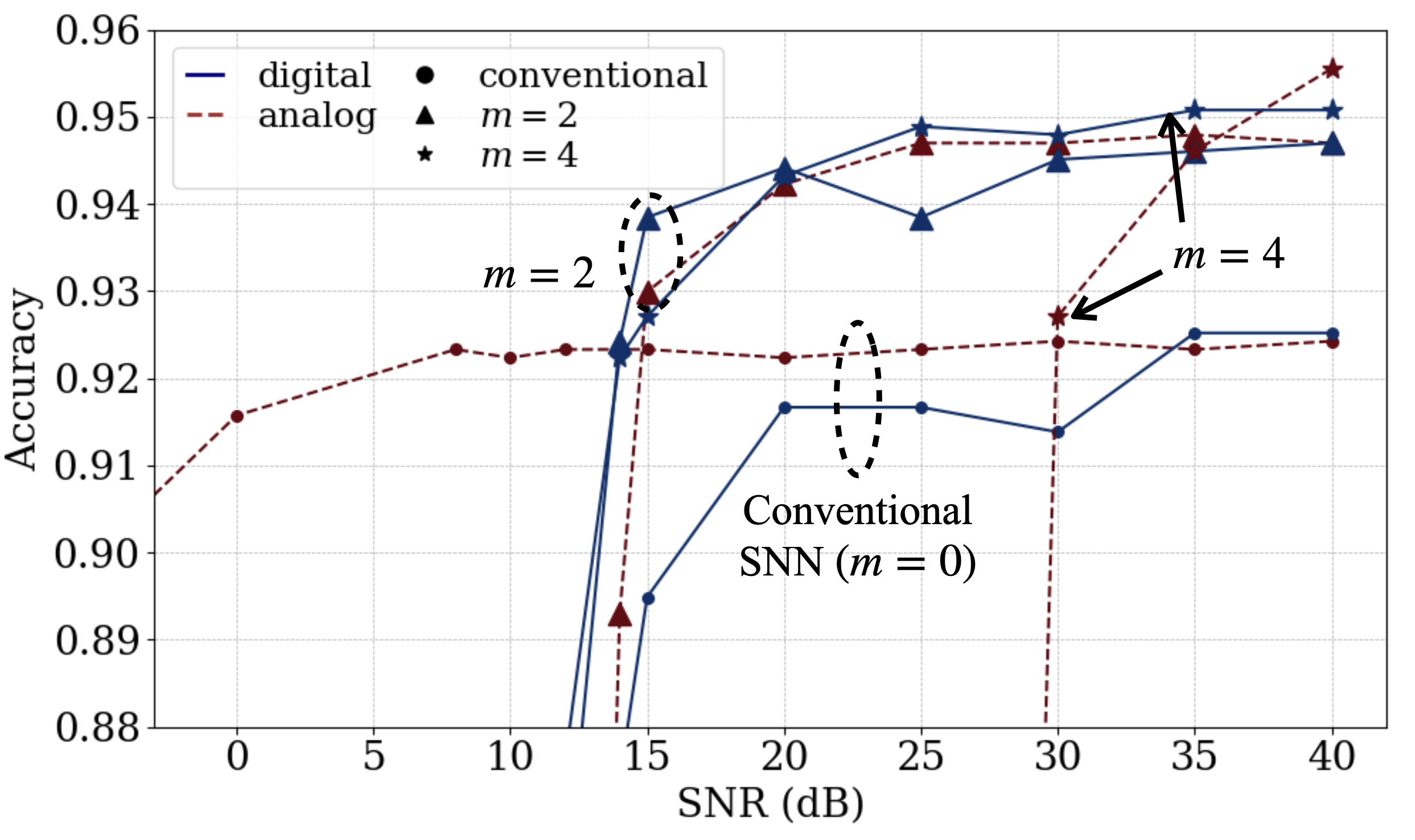}\label{fig:snr_exp_a}}
    \subfloat[Peak power constraint]{\includegraphics[width=0.5\columnwidth]{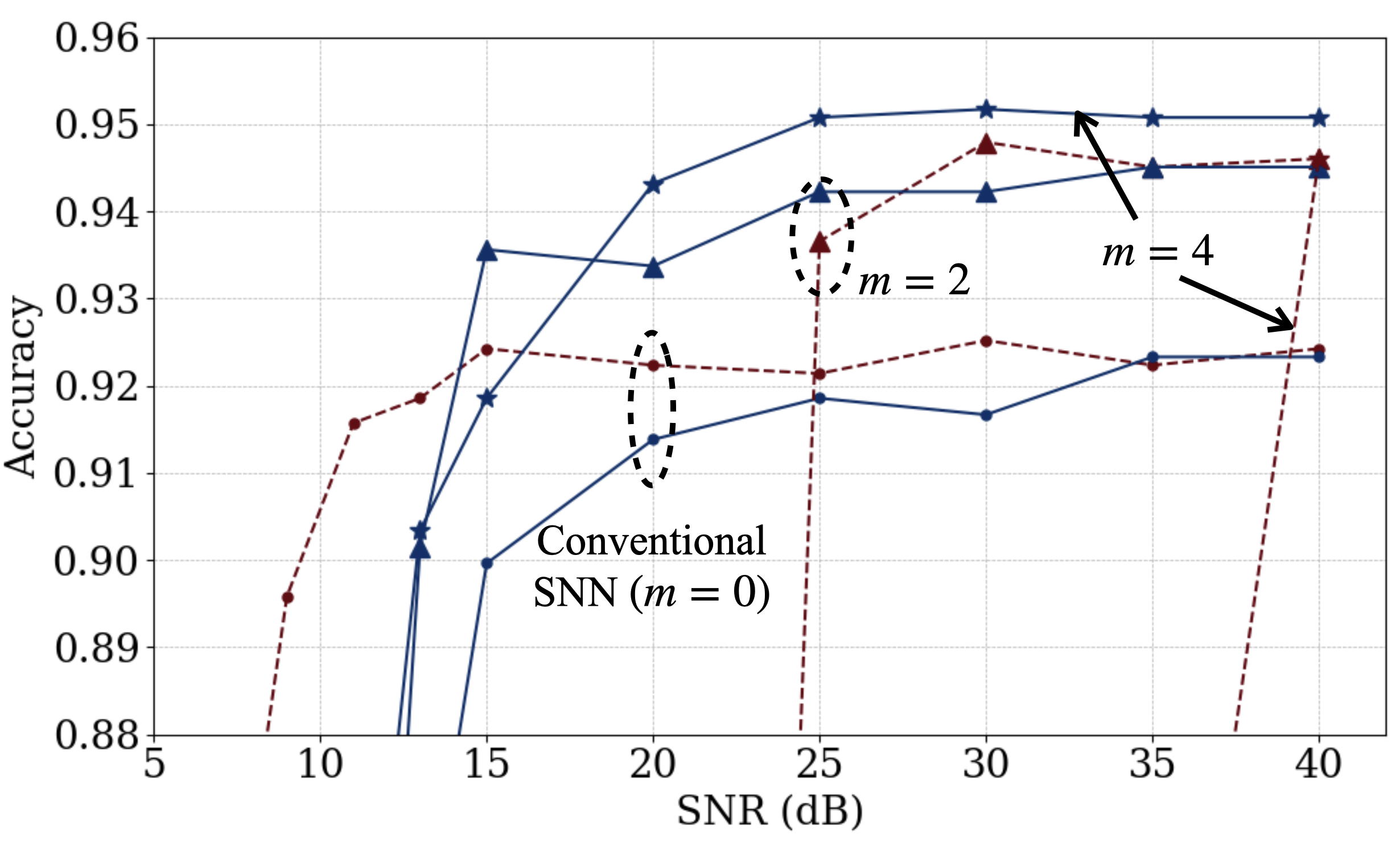}\label{fig:snr_exp_b}}
    \caption{Accuracy versus SNR for the neuromorphic wireless split computing architecture for analog and digital transmission schemes and: (a) per-block power constraint, and (b) peak power constraint (simulation, $T=4$).}
    \label{fig:snr_exp}
\end{figure*}

We first analyze the impact of the average SNR in Fig. \ref{fig:snr_exp} for both analog and digital modulation schemes with $N^{\mathrm{OFDM}}=2$ OFDM symbols under a per-block power constraint and a peak power constraint. A conventional SNN with analog modulation provides the best performance at lower SNR levels, especially under an average power constraint. In fact, a peak power constraint limits the power that can be used per spike, while an average power constraint makes it possible to allocate power by leveraging the sparsity of the spiking signals. At higher SNRs, it becomes essential to rely on muti-level spikes. In general, increasing the SNR calls for the selection of a larger value of $m$. This is aligned with the performance of the centralized implementation shown in Fig.~\ref{fig:acc_snr_noiseless}. 

\begin{figure}[t!]
    \centering
    \subfloat[Per-block power constraint]{\includegraphics[width=0.5\columnwidth]{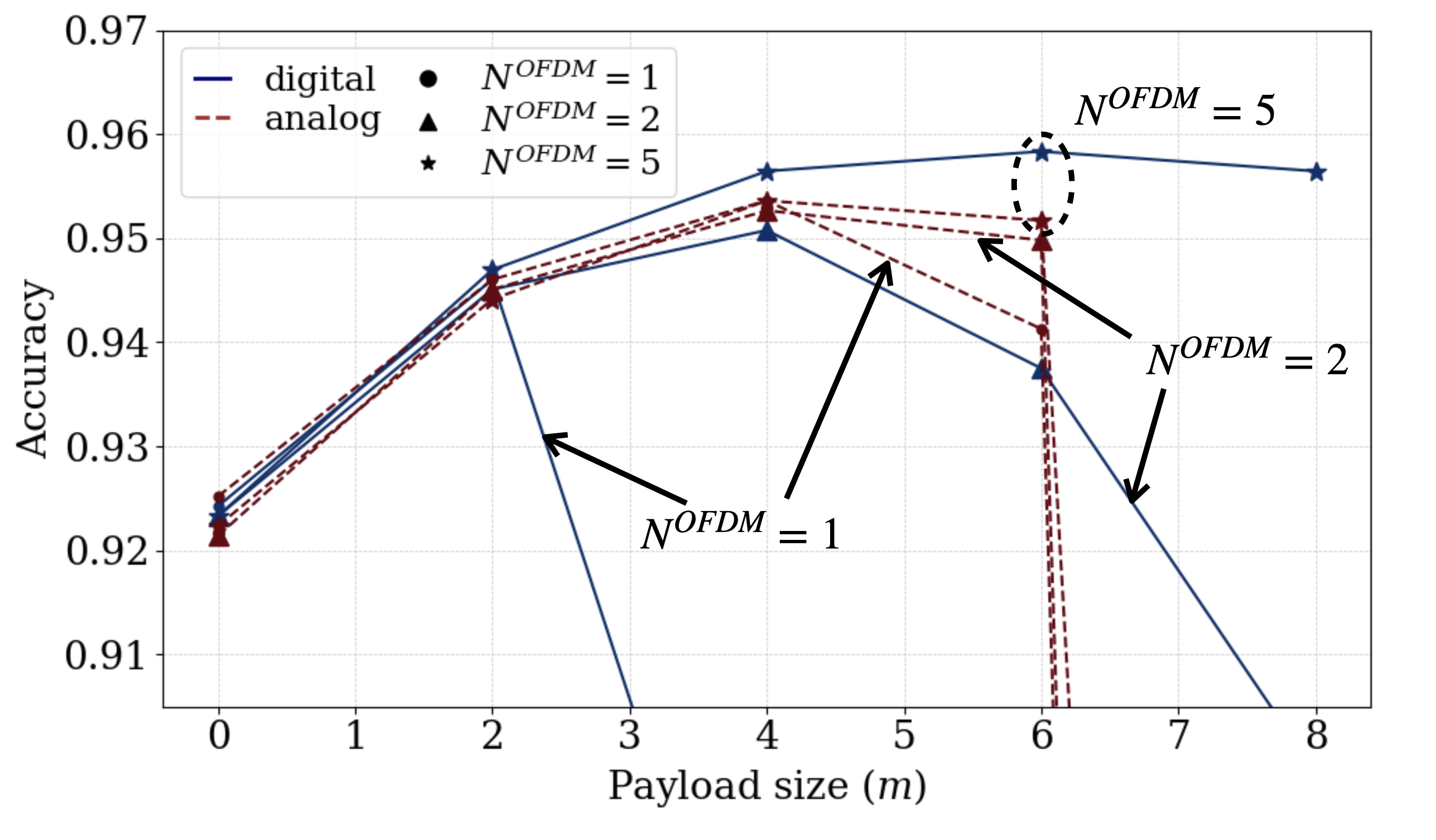}\label{fig:acc_ofdma_a}}
    \subfloat[Peak power constraint]{\includegraphics[width=0.5\columnwidth]{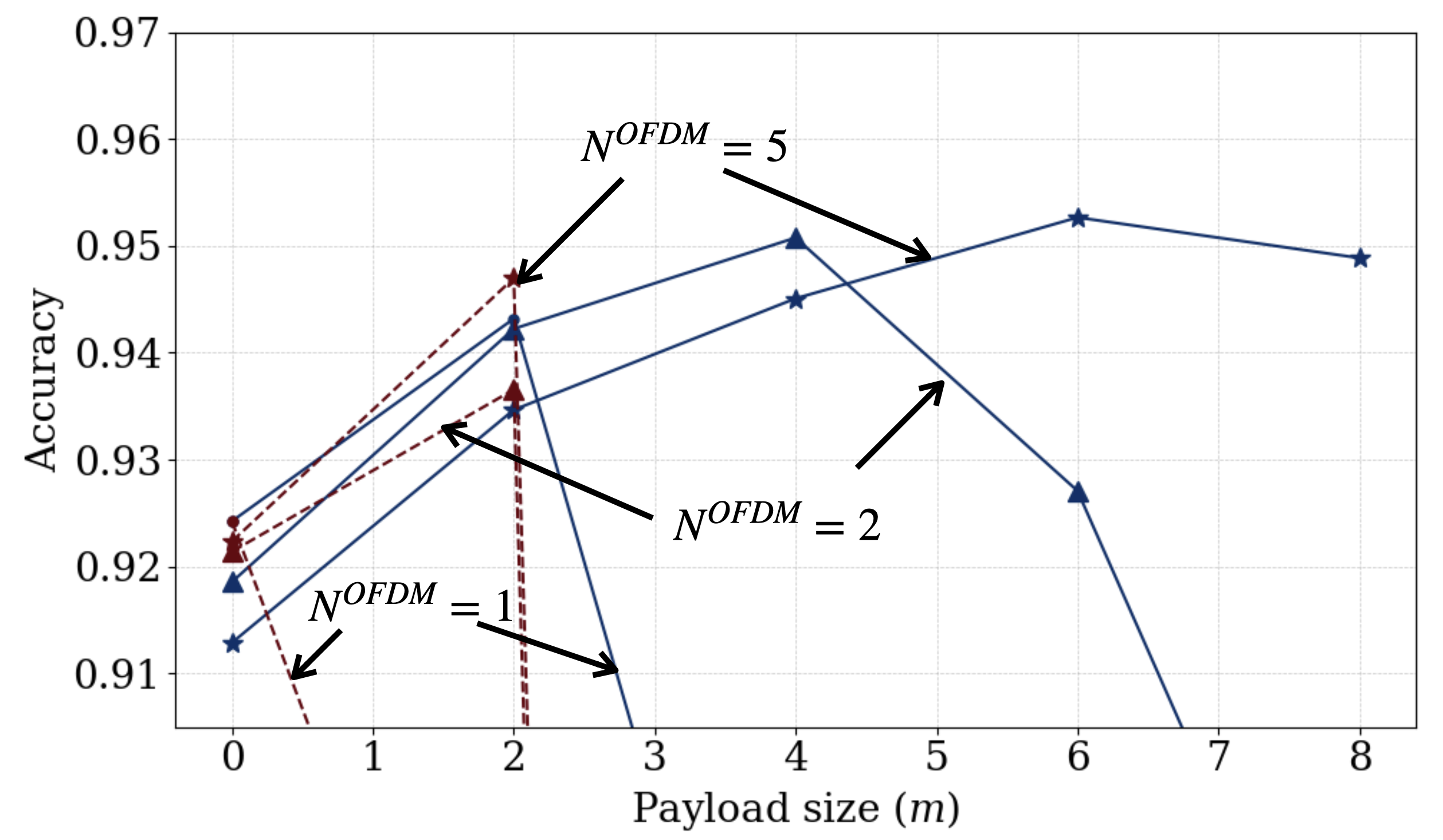}\label{fig:acc_ofdma_b}}
    \caption{Accuracy versus payload size $m$ for the neuromorphic wireless split computing architecture for analog and digital transmission schemes, using different number of OFDM symbols and: (a) per-block power constraint, and (b) peak power constraint (simulation, $T=4$).}
    \label{fig:acc_ofdma}
\end{figure}

\begin{figure}[!h]
    \centering
    \subfloat[Per-block power constraint]{\includegraphics[width=0.5\columnwidth]{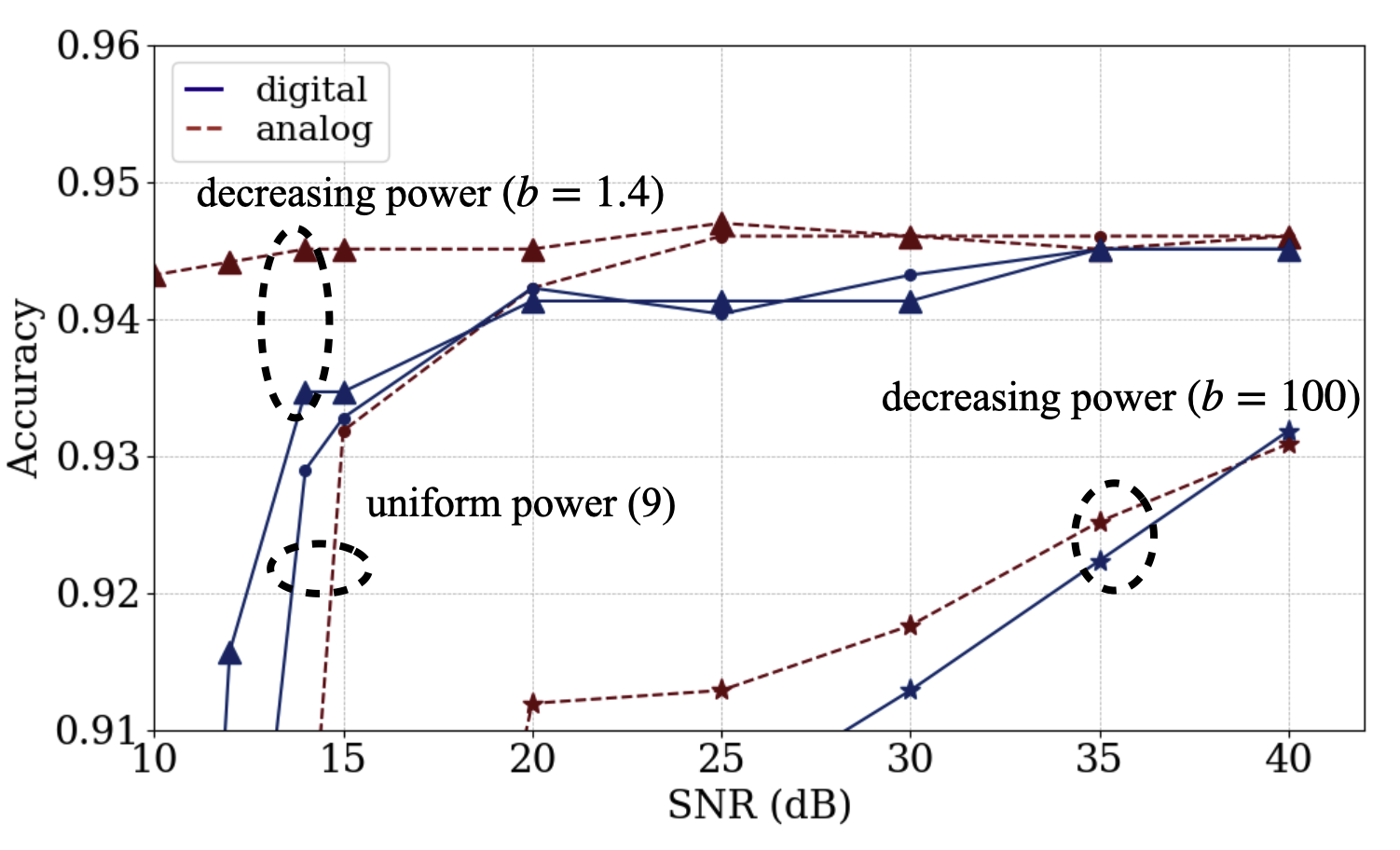}\label{fig:block_opt_power}} 
    \subfloat[Peak power constraint]{\includegraphics[width=0.5\columnwidth]{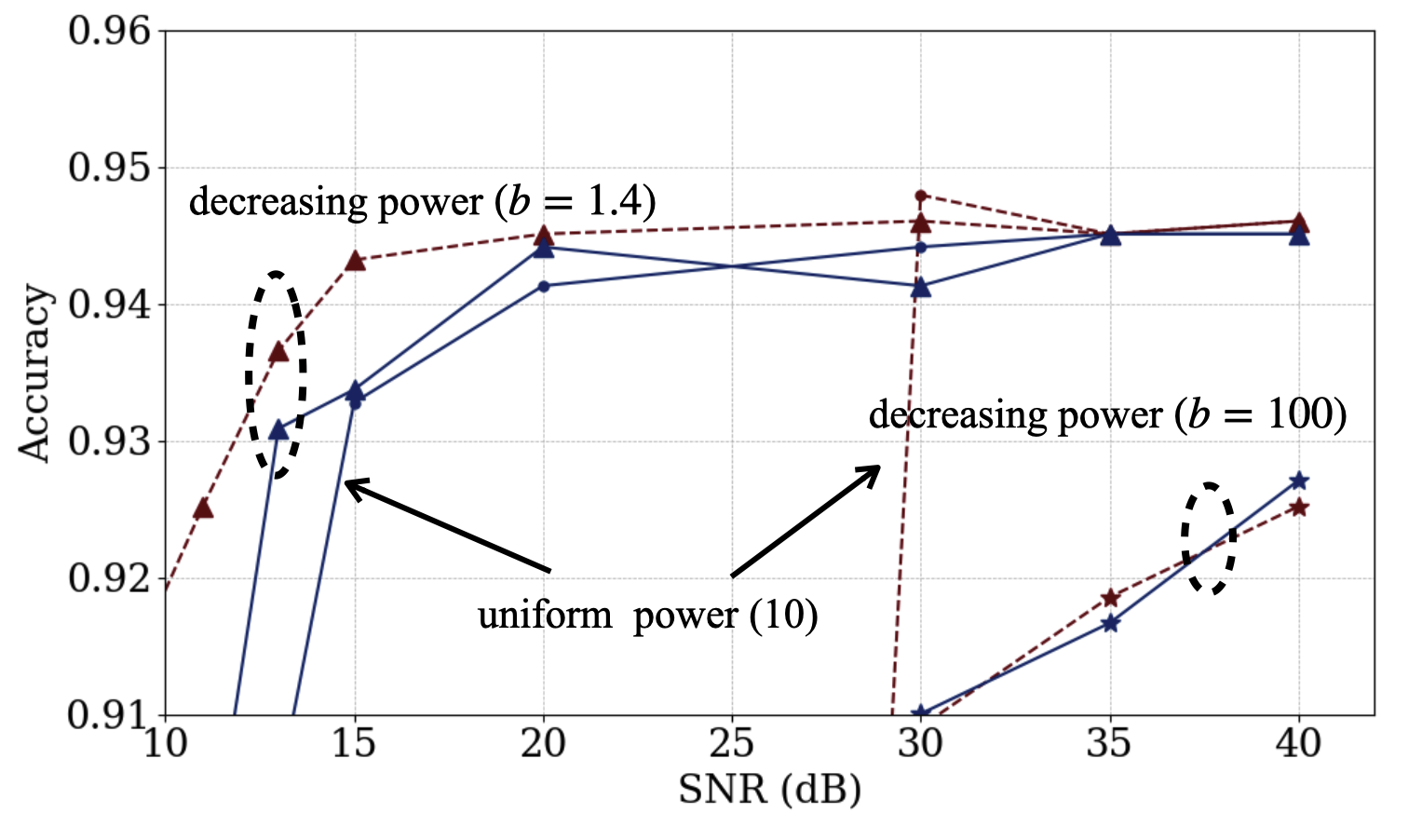}}\label{fig:peak_opt_power}
    \caption{Accuracy versus SNR for the neuromorphic wireless split computing architecture for analog and digital transmission schemes using: (a) per-block power constraint, and (b) peak power constraint (simulation, $m=2$ and $T=4$). For both types of power constraints, we consider time-uniform power allocation, as well as decreasing power allocation schemes as in \eqref{dypo} with different exponential decay rates $b$.}\label{fig:optimized}
\end{figure}

The optimal value of the payload size $m$ generally depends on the SNR, on the modulation schemes, and on the available spectrum, which is controlled by the number of OFDM symbols $N^{\rm OFDM}$. To elaborate on this, Fig. \ref{fig:acc_ofdma} presents the accuracy as a function of the bit width $m$ when the SNR is fixed at 25 dB, while varying also the number of OFDM symbols. The results indicate that for both modulation schemes, there is an optimal value of $m$ that strikes the best balance between increased inference accuracy and decreased transmission reliability caused by a larger value of $m$. 

In this regard, digital modulation is more sensitive to a decrease in spectral resources, showing a significant accuracy drop when $ m > 2 $ and $ N^{\mathrm{OFDM}} = 1 $, as limited resources force some of the spikes to be discarded. However, when the number of OFDM symbols is sufficiently large, such as $ N^{\mathrm{OFDM}} = 5 $, allowing most spikes to be transmitted, digital modulation can better capitalize on an increasing value of $m$. Under either power constraint, analog modulation also benefits from an increase in the number of OFDM symbols $ N^{\mathrm{OFDM}}$, becoming more robust through repetition coding.

Fig.~\ref{fig:optimized} presents the accuracy performance as a function of SNR for analog and digital transmission schemes under different power allocation strategies with $m=2$ and $T=4$. The results show that the decaying power allocation strategy in \eqref{dypo} can outperform fixed power allocation with a suitably chosen decay rate $b$, here $b=1.4$. This validates that errors occurring in earlier sensing slots can have a cascading effect, leading to larger degradation compared to those in later slots. However, an excessively large value of $b$, here $b=100$, can underperform uniform power allocation.

\begin{figure}[!h]
    \centering
    \subfloat[]{\includegraphics[width=0.5\columnwidth]{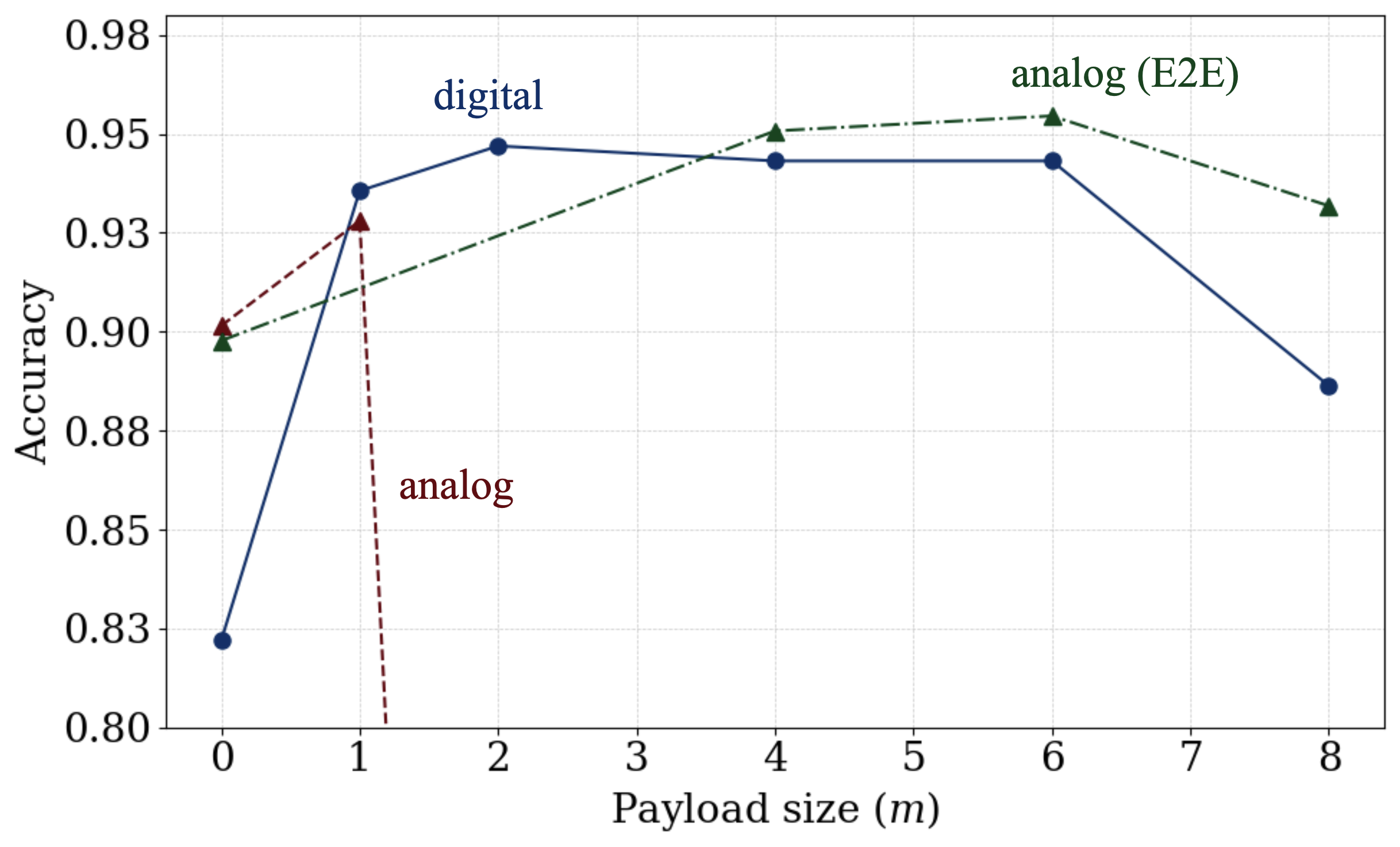}\label{fig:usrp_bit_width}} 
    \subfloat[]{\includegraphics[width=0.5\columnwidth]{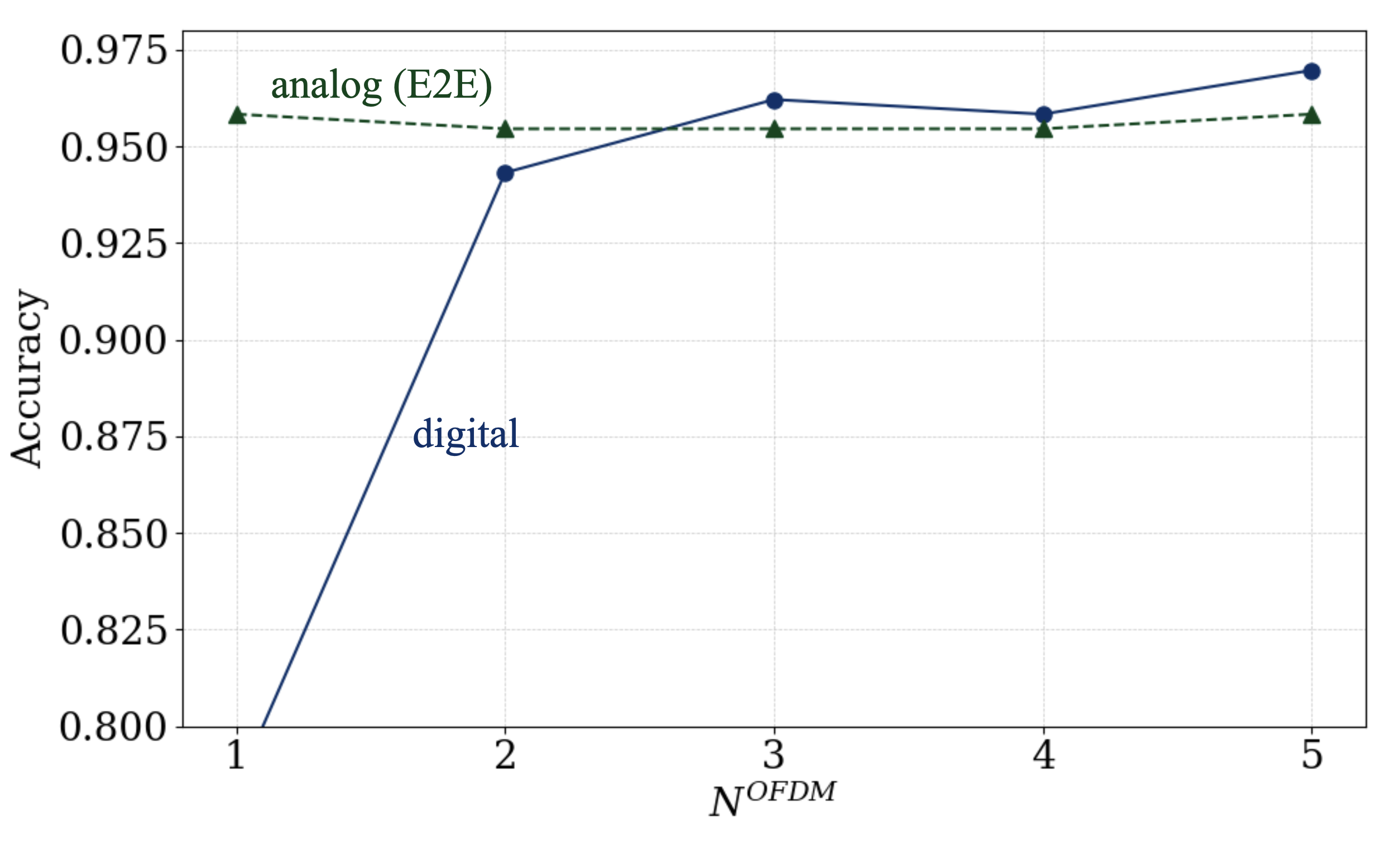}}\label{fig:usrp_ofdma}
    \caption{ Comparison between analog and digital transmission schemes for the neuromorphic wireless split computing architecture. (a) Accuracy versus payload size $m$ with $N^{\rm OFDM}=2$, and (b) accuracy versus number of OFDM symbol $N^{\rm OFDM}$ per sensing slot with payload size $m=6$ (USRP implementation, $T=4$). }\label{fig:usrp_epx}
\end{figure}

\subsubsection{Experimental Results}
We now turn to the results obtained from the real-world experiments with USRP radios. Fig. \ref{fig:usrp_epx}(a) presents the accuracy versus the number of bits $m$ for $N^{\rm OFDM}=2$. As discussed, for the analog implementation, we show the performance with  pre-trained models, as well as with end-to-end (E2E) fine-tuning with measured channels. As seen in the simulation in Fig.~\ref{fig:acc_ofdma}, there exists an optimal value of $m$ for all schemes. Furthermore, the figure highlights the importance of incorporating channel data in the optimization of a deployment that relies on analog transmission. 

Fig. \ref{fig:usrp_epx}(b) shows the accuracy versus the number of OFDM symbols, $N^{\rm OFDM}$, for $m=6$ bits, focusing on E2E fine-tuning for analog transmission. It is observed that digital modulation requires a sufficiently large number of OFDM symbols in order not to be limited by the accuracy degradation caused by spikes being dropped at the transmitter. In contrast, analog transmission can obtain the best performance even with only one OFDM symbol.

\section{Conclusions} \label{con}
In this paper, we have studied a neuromorphic wireless split computing architecture that leverages multi-level SNNs. Multi-level SNN models are known to achieve higher accuracy than conventional SNNs, especially in the presence of strict constraints on the sensing period. However, in a split computing system, these gains may be offset by the challenges of exchanging multi-level spikes between the SNN models deployed across two separate devices. To address this problem, we have developed digital and analog modulation schemes optimized for an OFDM radio interface. Simulations and experiments with software-defined radios have accordingly revealed optimal configurations in terms of the size of the spike payload for both analog and digital transmission schemes. Analog transmission was seen to perform better at lower SNR levels and for smaller payload sizes, while digital transmission was seen to be more effective at higher SNR levels and for larger payloads. Finally, experimental results have demonstrated the need for channel-specific fine-tuning of the SNN models for analog transmission. Future work may consider extensions to multi-terminal settings \cite{chen2023neuromorphic}, the use of UWB for short-range low-power communications, and the problem of dynamic power optimization across sensing slots, e.g., via reinforcement learning.

\small{
\bibliographystyle{IEEEtran}
\bibliography{references}}

\end{document}